\documentclass[a4paper,11pt,numbers]{elsarticle}

\usepackage{graphicx}
\usepackage[fleqn]{amsmath}
\usepackage{amsfonts}
\usepackage{amssymb}
\usepackage{float}
\usepackage[dvips]{epsfig}
\usepackage{epstopdf}
\usepackage{rotating}
\usepackage{epigraph}
\usepackage{diagbox}

\usepackage{setspace} % Lots of packages that Tim likes :-)
\usepackage{lineno}   %,hyperref}
\modulolinenumbers[5]

\usepackage{microtype}
\usepackage{url}
\usepackage{subfiles}
\usepackage{caption}
\usepackage{subcaption}
\usepackage{parskip}
\usepackage[section]{placeins}
\usepackage{xcolor}
\usepackage{framed}
\colorlet{shadecolor}{blue!20}
\usepackage{tabulary}
\usepackage[hidelinks]{hyperref}
\usepackage[capitalize,nameinlink]{cleveref}
\usepackage{flexisym}

\usepackage{algorithm}
\usepackage{algpseudocode}

\usepackage{amsmath,amsthm,amssymb}
\usepackage{calrsfs}
\DeclareMathAlphabet{\pazocal}{OMS}{zplm}{m}{n}

\usepackage{multirow, bigdelim}

\usepackage{tikz}
\usetikzlibrary{decorations.pathmorphing}
\usetikzlibrary{decorations.markings}
\usetikzlibrary{arrows.meta,bending}
\usetikzlibrary{arrows.meta}
\usetikzlibrary{positioning}
\usepackage{bm}

\usepackage{svg}

\usepackage{adjustbox}

\usepackage{standalone}

\graphicspath{{./Figures/}}

\journal{}

\begin{document}
\journal{Mechanical Systems and Signal Processing}

\begin{frontmatter}

\title{Towards a population-informed approach to the definition of data-driven models for structural dynamics}

\author[]{G.\ Tsialiamanis\footnote{Corresponding Author: George Tsialiamanis (g.tsialiamanis@sheffield.ac.uk)}}
\author[]{N.\ Dervilis}
\author[]{D.J.\ Wagg}
\author[]{K.\ Worden}
\address{Dynamics Research Group, Department of Mechanical Engineering, University of Sheffield \\ Mappin Street, Sheffield S1 3JD}

\begin{abstract}
Machine learning has affected the way in which many phenomena for various domains are modelled, one of these domains being that of structural dynamics. However, because machine-learning algorithms are problem-specific, they often fail to perform efficiently in cases of data scarcity. To deal with such issues, combination of physics-based approaches and machine learning algorithms have been developed. Although such methods are effective, they also require the analyser's understanding of the underlying physics of the problem. The current work is aimed at motivating the use of models which learn such relationships from a population of phenomena, whose underlying physics are similar. The development of such models is motivated by the way that physics-based models, and more specifically \textit{finite element} models, work. Such models are considered transferrable, explainable and trustworthy, attributes which are not trivially imposed or achieved for machine-learning models. For this reason, machine-learning approaches are less trusted by industry and often considered more difficult to form \textit{validated} models. To achieve such data-driven models, a \textit{population-based} scheme is followed here and two different machine-learning algorithms from the \textit{meta-learning} domain are used. The two algorithms are the \textit{model-agnostic meta-learning} (MAML) algorithm and the \textit{conditional neural processes} (CNP) model. The two approaches have been developed to perform within a population of tasks and, herein, they are tested on a simulated dataset of a population of structures, with data available from a small subset of the population. Such situations are considered to be similar to having data available from existing structures or structures in a laboratory environment or even from a model and needing to model a new structure with only a few available data samples. The algorithms seem to perform as intended and outperform a traditional machine-learning algorithm at approximating the quantities of interest. Moreover, they exhibit behaviour similar to traditional machine learning algorithms (e.g. neural networks or Gaussian processes), concerning their performance as a function of the available structures in the training population, i.e.\ the more training structures, the better and more robustly the algorithms learn the underlying relationships.\end{abstract}

	
\end{frontmatter}

\graphicspath{{Figures/}}

\section{Introduction}
\label{sec:introduction}

Modelling of structural dynamics has been extensively studied throughout the years. The need to study how structures behave and react to the various conditions of their environment is motivated by the need to create safe and long-lasting infrastructure. This need has led researchers to try to understand the physics of various materials and of structural members made from such materials. Such studies led to creation of mathematical models, founded on the understanding of the physics of the various components of the structures. 

Numerous methods have been developed to analyse and predict the behaviour of structures, but, arguably, the most common and successful means of structural analysis is the \textit{finite element} (FE) method \cite{bathe2006finite}. The method has been widely successful because of its generality and accuracy, especially for static problems. Although the FE method was developed many years in the past, the advance of computational resources \cite{huthwaite2014accelerated} have allowed the modern use of ever more detailed and complicated FE models, maintaining the method as the dominant structural-analysis strategy.

Naturally, models are used to simulate structures before fabrication and after implementation. For the former case, one typically has no available data in order to define the parameters of the model. The selection of the parameters is performed using engineering intuition and available data from experiments or from existing structures, built with similar materials and comprising similar structural members. For the latter case however, the analyst may have acquired some data from the existing structure and may seek to tune the parameters of a model, so that the predictions of the model align with the recorded data. The knowledge acquired from such a model calibration procedure is in the form of values, or distributions of values, for the tunable parameters of the models and can be exploited in similar tasks in the future.

In recent years, because of the vast development of data-driven modelling methods for many disciplines and because of how conveniently such methods fit the framework of using data acquired from a structure in order to model it, data-driven structural modelling has emerged as a viable approach. The field which has contributed most, is that of \textit{machine learning} \cite{Bishop, Bishop2, Murphy:2012:MLP:2380985, goodfellow2016deep}, and it has been extensively used for the purposes of structural dynamics \cite{cross2022physics,tygesen2019state}, and quite often for \textit{structural health monitoring} (SHM) \cite{Farrar}. The success of the application of such methods lies in the fact that they learn the underlying relationships of the modelled quantities from data and do not require extensive physical understanding of these relationships by the creator of the model. Data-driven methods are also able to effectively simulate physics of different domains and without the restriction of time and space scales.

However, the ability of data-driven models to learn without knowledge of the underlying physics, comes with a shortcoming - the need for data. Depending on the problem and the algorithm that is used, the amount of available data and their quality might be restrictive in terms of performing appropriate inference. In extreme cases data may not be available at all, making the use of such algorithms difficult or even infeasible. In the field of structural dynamics, this problem is extensive \cite{gardner2020application}. Data from structures are not always available, need specialised equipment to be acquired and in many cases, the environmental conditions during the acquisition of the data may not be the same as the conditions under which one seeks to model the structure. The latter problem is evident when one performs experiments in a laboratory and uses these data to model a structure in the field.

Physics-based models do not suffer very often from lack of such data. The reduced need for data is balanced by the imposition of physics into the model. Assuming that the physics are accurate to a prescribed level, the model would be expected to be able to generalise well, even if it is calibrated using only a small number of data points \cite{raissi2019physics}. However, the physics are not the only factor that allow these models to generalise that well. An aspect of such models is the convenience of knowledge transfer between members in a population of structures. As mentioned, studying structures started by studying materials and simple components in a laboratory. Subsequently, the acquired knowledge is used to model real and complicated structures based on the behaviour observed in the laboratory. The results of calibrating a model, e.g.\ a FE model, can provide a prior belief, which could assist in modelling a structure comprising the same materials and same type of members. 

Apart from the reduced need for data, a convenient characteristic of physics-based models - such as FE models - is that they are preferred because of their \textit{explainability}. The models have been used for several decades and details about their functionality are known either by analyses that have been performed, or by hands-on experience of the users. This explainability also comes from the fact that their tunable parameters are quite specific and the effect of variations in these parameters can be systematically studied. A large contrast to the physical trainable parameters is identified if one compares them with the trainable parameters of neural networks, which quite often have no physical meaning and their variations cannot be correlated to specific physically-meaningful effects on the behaviour of the model.

Motivated by this functionality of physics-based models, the current work is focussed on the use of data-driven models, which are informed by populations of structures in order to boost their performance, increase their trustworthiness and impose a sense of explainability. Similar population-based approaches have very recently been followed for SHM problems \cite{PBSHMMSSP1, PBSHMMSSP2, PBSHMMSSP3, PBSHMMSSP4, bull2022knowledge}. The approach proposed here-in is to define data-driven models which are able to adapt to different members of the population, in a similar manner to how corresponding physics-based models are used in a population. In Section \ref{sec:PB_modelling}, the motivation to create such data-driven models is given, as well as the formulation of such models and the potential benefits of using a population-based approach. In Section \ref{sec:meta_learning_PB_modelling}, an introduction to \textit{meta-learning} \cite{vanschoren2018meta, hospedales2020meta}, is given, together with an explanation of its connection with population-based modelling and two algorithms are presented as an attempt to define population-informed models which extract knowledge from the population without the need for the analyser to impose prior knowledge. In Section \ref{sec:applications}, applications are presented to illustrate the potential of the algorithms in comparison to a traditional machine-learning method, which is trained without exploiting any population knowledge. Finally, in Section \ref{sec:conclusions}, conclusions are drawn and future work is discussed.

\section{Population-based modelling}
\label{sec:PB_modelling}

A general framework for physics-based modelling follows a mathematical formulation of the physics of some phenomenon and the use of the model to predict responses in hypothetical scenarios. For structural dynamics, the objects under study are structures or structural members and often, the inputs to the models are environmental and operational conditions or loadings imposed by the environment. One example of such a situation is a wind-turbine and the corresponding environmental condition might be the ambient temperature and a potential excitation signal might be the time-history of the wind speed. 

The variables affecting the output of the physics-based model can be separated into two categories. The first category comprises the variables $\bm{x}$ that externally affect the behaviour of the structure. These variables may be environmental variables, such as temperature and humidity or some external loading, such as an excitation force, an earthquake excitation etc. The second category refers to the parameters of the model $\bm{c}$, which describe quantitatively the physics of the structure which is modelled. Such parameters are often the Young's modulus of the material, the Poisson's ratio etc. The latter category contains the parameters of a model, which are tuned during the calibration procedure. The quantity of interest $\bm{y}$ is then given by,
\begin{equation}
    \bm{y}(\bm{x}) = f^{p}(\bm{x}; \bm{c})
    \label{eq:physics_based_1}
\end{equation}
where $f^{p}$ is a physics-based model and the values of the parameters $\bm{c}$ are chosen by the framer or the user of the model (at least their nominal values are chosen, since the actual values might vary because of the effect of environmental parameters). Naturally, the above formulation is subjected to uncertainty. The uncertainty is often separated into two categories, the \textit{epistemic uncertainty} and the \textit{aleatory uncertainty}. The first category mainly refers to inconsistencies between the formulation of the model $f^{p}$ and the phenomenon which it is used to model. The second category refers to inherently random quantities or events. Throughout this paper, vectors are represented by bold symbols, while scalar variables have regular symbols.

A calibration procedure according to observations of the quantities of interest $\bm{y}$ yields a set of optimal parameters for the values of the tunable parameters $\bm{c}$ of the model. The convenience of physics-based models lies in the fact that the tunable parameters refer to properties of the materials and the mathematical formulation of the physics is assumed to be common for structures made by similar materials. As a result, to extend the modelling framework in a population scheme, the modelled quantities $\bm{y}^{i}$ of the $i$\textsuperscript{th} structure are given by,
\begin{equation}
    \label{eq:physics_based_geometry}
    \bm{y}^{i}(\bm{x}) = f^{p}(\bm{x}; \bm{c}^{i}, \pazocal{G}^{i})
\end{equation}
where $\bm{c}^{i}$ are the parameters of the $i$\textsuperscript{th} structure and $\pazocal{G}^{i}$ is the geometry of the $i$\textsuperscript{th} structure. The geometry of the structure, which is incorporated in the physics-based model, is another part of the physics that the users incorporate into the model. In the case of FE models, the geometry refers to the number and coordinates of the nodes of the model and the boundary conditions. Transferring knowledge within a population is manual, but formulated and facilitated by the mathematical formulation, as the values of the material parameters $\bm{c}$ are transferred from one model to the other and the geometry $\pazocal{G}$ is adapted to reflect the new structure, which is to be modelled.

The corresponding framework for structural modelling using data-driven models is similar. A set of quantities of interest $\bm{y}$ is modelled as a function of some input variables $\bm{x}$. In the case of modelling a single structure, the data-driven model $f^{d}$ has trainable parameters $\bm{\theta}$ and the equation of the predictions is given by,
\begin{equation}
    \label{eq:data_driven_1}
    \bm{y}(\bm{x}) = f^{d}(\bm{x}; \bm{\theta})
\end{equation}

However, in a population-based framework, the approach to transfer knowledge is not the same as in physics-based models. The trainable parameters $\bm{\theta}$ cannot be trivially transferred or transformed, as in the case of the parameters $\bm{c}$ and the geometry $\pazocal{G}$, in order to be used for another structure with geometry $\pazocal{G}^{*}$ within a population.

Under such a framework, attempts at knowledge transfer have been made for SHM in \cite{PBSHMMSSP1, PBSHMMSSP2, PBSHMMSSP3, PBSHMMSSP4}. The first and simplest approach is based on structures in \textit{homogeneous} populations \cite{PBSHMMSSP1}. In that case, the structures are considered similar enough, so that the same model $f^{d}$ shall perform satisfactorily for all of them. To follow such an approach, one needs to find the set of best fitting parameters $\bm{\theta}$, by solving the optimisation problem given by,
\begin{equation}
    \label{eq:optimisation}
    \hat{\bm{\theta}} = \min_{\bm{\theta}} \pazocal{L}(\bm{y}^{i}(\bm{x}), f^{d}(\bm{x}; \bm{\theta})) \quad i=1, 2..., N
\end{equation}
where $\pazocal{L}$ is some objective function expressing the distance between the predictions of $f^{d}(\bm{x}; \bm{\theta})$ and the observations $\bm{y}^{i}(\bm{x})$ of the $i$\textsuperscript{th} structure, $N$ is the number of available structures and $\hat{\bm{\theta}}$ captures the optimal values of the parameters $\bm{\theta}$, which minimise $\pazocal{L}$. In the equation above, the observations of all N structures are included in the optimisation procedure, making the model common for all the population or, as such models are called in \cite{PBSHMMSSP1}, a \textit{form}. Such a model can also be a probabilistic model, yielding a probability density function $p(\hat{\bm{\theta}})$ instead of single point predictions, i.e.\ $p(\bm{y}|\bm{x}) = f^{d}(\bm{x}; \hat{\bm{\theta}})$.

Structures with significant differences form \textit{heterogeneous} populations, making knowledge transfer more difficult. Various approaches have been followed to deal with such problems. In \cite{PBSHMMSSP3, gardner2022application, gardner2022population}, \textit{domain adaptation} is exploited to perform such knowledge transfer. The problems presented refer to classification of damage on structures with significant differences and how a damage classifier could be transferred and adapted to different structures. 

The data-driven methods described so far do not take into account the geometry and do not seek to exploit any other physical knowledge about the structures. To consider the geometry in the inference process, in \cite{PBSHMMSSP2, PBSHMMSSP4}, a graph representation of the structures is used. The representation is performed by breaking down the structures into \textit{irreducible elements} (IEs). Each IE represents a structural element and a graph is constructed by connecting the IEs with each other according to the connectivity of the structural elements. The graphs are attributed according to the properties of the structural members and are used either to classify structures according to their similarities \cite{PBSHMMSSP2, brennan2022similarity} or to perform inference within a population comprising structures created by a set of members of the same type. The geometry is part of the input, as in the case of physics-based models, like FE models, partly compensating for the lack of a physical formulation. The output of such graph-based models $f^{g}$ is in this case given by,
\begin{equation}
    \bm{y}^{i} = f^{g}(\bm{x}, \pazocal{G}^{i}; \bm{\theta})
\end{equation}
where $\bm{x}$ is some external input which affects the behaviour of the structure and $\pazocal{G}^{i}$ is the geometry, represented via a graph, of the $i$\textsuperscript{th} structure.

In order to compare the physics-based and data-driven approaches, one should compare equations (\ref{eq:physics_based_1}) and (\ref{eq:data_driven_1}). Doing so, a parallelism is observed. In both cases, the modelled quantities $\bm{y}$ are a function of some inputs-to-the-model $\bm{x}$ and some parameters of the model, either $\bm{c}$ and $\pazocal{G}$ in equations (\ref{eq:physics_based_1}) and (\ref{eq:physics_based_geometry}) or $\bm{\theta}$ in equation (\ref{eq:data_driven_1}). The major difference between physics-based and data-driven models can be summarised in the difference between the two sets of parameters $\bm{c}$ and $\bm{\theta}$. 

On the one hand, the parameters $\bm{c}$ describe the physics of the structure; they vary within an interval of potential values according to the phenomena they describe and are chosen by a user so that they best reflect the physics of the object modelled. Their values are also explainable and \textit{meaningful}; for example, a material with higher value of Young's modulus than some other material is a more stiff material. During a model calibration procedure, a set of model parameters $\bm{c}$ is sought, which will sufficiently fit the acquired data from some structure. The resulting set of parameters has a strong physical meaning, since the model is bound by the physics imposed by its mathematical formulation. In some cases, calibration may be performed manually using a trial and error technique by an experienced user. Hence, one is confident that, using the same or similar parameters for similar structures, will result in good accuracy for predictions regarding the new structure.

On the other hand, the parameters $\bm{\theta}$ are selected via the optimisation described by equation (\ref{eq:optimisation}) and usually have no physical meaning. Taking also into account that \textit{black-box} models are often overparametrised functions (e.g.\ neural networks); the set of parameters, which best describes a relationship, is not necessarily unique. The latter issue of non-uniqueness creates problems when such models are used in a population-based framework. As in the case of physics-based models, it would be convenient if the population could be described by varying parameters and if smooth alterations of these parameters would correspond to smooth alterations of $\bm{\theta}$. However, this is not the case, since some parameters $\bm{\theta}_{i}$ may describe sufficiently the physics for some structure $i$, but for a different structure with slightly different characteristics, the optimal set of parameters may not even be in the neighbourhood of $\bm{\theta}_{i}$. For the case of physics-based models, the set of all plausible values of $\bm{c}$ defines all the potential forms of the model and describes a population parametrised by $\bm{c}$ and its plausible values. In contrast, a data-driven model $f^{g}(\bm{x}; \bm{\theta})$ does not define a meaningful set of mappings for every value of $\bm{\theta}$, since for the majority of the values of $\bm{\theta}$, the model may not resemble any plausible reality. 

Trying to create a data-driven model which shall be able to perform within the population and not for a single structure, one could attempt to create a model that takes as inputs the parameters $\bm{c}$, which describe every structure of the population. The modelled quantities $\bm{y}$ will then be given by,
    \begin{equation}
    \label{eq:data_driven_2}
    \bm{y}(\bm{x}, \bm{c}) = f^{d}(\bm{x}, \bm{c}; \bm{\theta})
\end{equation}
This approach outweighs the use of a physics-based model as in equation (\ref{eq:physics_based_1}), in the case of existence of epistemic uncertainty for the physics-based model; i.e.\ when one is not confident that the physical formulation of $f^{p}$ does sufficiently describe the underlying physics of the problem.

A first problem with this approach is that the available data for the population may often be imbalanced, in the sense that more data may be available for some structure $i$, and the available data for some other structure $j$ may be restricted. In some cases, this imbalance may not be a major issue, but in general, for structural dynamics, datasets tend to be quite sparse. Commonly, for a small number of structures, plenty of data are recorded, while for other structures only a few samples are available. The imbalance may come from the fact that some structures are in a laboratory environment and tested extensively or accurate models may exist for structures, which are trustworthy and allow testing a structure under various ‘what-if’ scenarios. Furthermore, if the data of the available structures refer to a small subregion of the parameter space of the structures, the model $f^{d}$ shall essentially be called on to \textit{extrapolate} for other subregions of parameters, a functionality that most machine-learning algorithms lack. A type of machine learning models that exhibit some extrapolation capabilities are the \textit{physics-informed neural networks} (PINNs) \cite{raissi2019physics}.

The second major problem is the \textit{parametrisation} of the structures. For FE models the geometry $\pazocal{G}$ serves as a parametrisation technique; as described by equation (\ref{eq:physics_based_geometry}). For data-driven models and simple cases, such as a population of beams or cantilevers, the parametrisation via the use of the dimensions of the structures is straightforward. However, in other cases, it can be extremely complicated and even infeasible. For example, which parameter vector $\bm{c}$ should be assigned to the wing of a commercial aircraft and which to the wing of a military aircraft? Both wings are intuitively placed in the same population of structures, but that does not mean that defining a set of parameters which characterise the population is a trivial task. To deal with such issues, as mentioned, in \cite{PBSHMMSSP2}, a breakdown of structures into irreducible elements (IEs), and a conversion of structures into graphs is presented. In \cite{PBSHMMSSP4}, an attempt is made to use such graphs in order to perform inference for the population. However, the algorithm presented in \cite{PBSHMMSSP4} requires large amounts of data and an automatic and holistic way to transform structures into graphs. Steps towards an objective transformation of structures into graphs have been recently made in \cite{brennan2022similarity}.

To deal with the aforementioned problems during the definition of a data-driven problem, a population-based approach is proposed in the current work. The population-based approach, as mentioned, is motivated by PBSHM. The use of a population of structures as a means to train data-driven models is expected to add useful features to these models, that they lack compared to physics-based options. Such features are the ability to be trained and to perform with a few available data samples from a testing structure. Moreover, because the \textit{validation} of models is becoming increasingly important \cite{worden2020digital}, the population-informed approach aims at creating models which, being validated for some structures of the population, are more trustworthy than data-driven models which are not informed somehow by the population. The described approach of using data-driven models may also result in quite explainable models. The explainability is encouraged by the fact that the model shall follow the rules of the population for predictions outside the domain of available data for a testing structure. In contrast completely black-box models may behave almost randomly outside their training domain.

% To deal with the described issues, in the current work, meta-learning is employed. The method used is that of \textit{model-agnostic meta-learning} (MAML) \cite{finn2017model}. The method is used to create neural network models that are able to quickly and with a few data adapt to new tasks. The tasks, in the current framework, are structures of some population. The model seems to be efficiently learning the relationship of some quantity of interest of a set of structures from a population as a function of some environmental variable. Moreover, the algorithm bypasses the issue of mapping structures onto a set of parameters $\bm{c}$.

\section{Two approaches for population-informed neural network models of structures}
\label{sec:meta_learning_PB_modelling}

\subsection{Meta-learning for population-based modelling}

The aim of the current work, is to create a data-driven model - more specifically a neural network - which shall be used within a population of structures, in a similar manner that a physics-based model would be used. The similarity which is sought, is that between equations (\ref{eq:physics_based_1}) and (\ref{eq:data_driven_1}). As mentioned, a physics-based model, following equation (\ref{eq:physics_based_1}) with varying parameters $\bm{c}$, is able to model a potential population of structures which is parametrised by $\bm{c}$. As a first attempt towards this direction it would be convenient and desirable to have a similar data-driven model as of equation (\ref{eq:data_driven_1}), whose parameters $\bm{\theta}$ would vary and their variation would be in order to model a different member of the population. The space of parameters $\bm{\theta}$ is essentially $\mathbb{R}^{n_{\bm{\theta}}}$, where $n_{\bm{\theta}}$ is the number of trainable parameters of the neural network model $f^{d}$. A way to force variation of the parameters to explicitly reflect the various structures of the population is to allow the parameters to vary only within a manifold of $\mathbb{R}^{n_{\bm{\theta}}}$. Having the parameters ``trapped'' on this manifold, creates a model $f^{d}$ whose parameter variations have acquired some physical meaning, i.e. each point of the manifold describes a different structure of the population (which would otherwise be described by $\bm{c}$).

Regarding the types of populations to which one could apply such a framework, the homogeneity of the population should be taken into account. For quite similar structures, such an approach would be more efficient, similarly to physics-based modelling, where for a homogeneous population, variations of a small subset of parameters (e.g. material-properties parameters) would suffice to model the population. For more heterogeneous populations in the physics-based modelling framework, the geometry of the model might have to change to account for the different structures. The homogeneity of the populations in the current work plays an important role, however, homogeneity in this case should refer to homogeneity of the underlying functions that are modelled. Intuitively, one would expect that homogeneity in the population would lead to homogeneity in the modelled functions. Therefore, it would be safer to follow an approach like the one described in \cite{PBSHMMSSP2} to define a sufficiently-homogeneous population to apply the current methods.

In a structural context, the desired behaviour of the algorithm can be described here via an example. Consider a simple simulated lumped-mass system, as shown in Figure \ref{fig:mass_spring_example}. The mass-spring system is defined by its structural parameters: the stiffness $k$, the damping coefficient $c$ and its mass $m$. The physics of the system are dictated by the equation of motion,
\begin{equation}
    M\bm{\Ddot{y}} + C\bm{\Dot{y}} + K\bm{{y}} = \bm{x}(t)
\end{equation}
where $M$, $C$, $K$ are the mass, damping and stiffness matrices of the system, $\bm{\Ddot{y}}$, $\bm{\Dot{y}}$, $\bm{{y}}$ are the vectors of accelerations, velocities and displacements of the degrees of freedom of the system and $\bm{x}(t)$ is the vector containing the external forces applied to the degrees of freedom of the system as a function of time. Since the system, as mentioned, belongs to a population of similar systems, many such systems are considered here by varying the stiffness parameter $k$. The system may also be considered to have varying structural parameters as a function of some environmental parameters $\bm{e}$; e.g.\ temperature, humidity etc. In the current case, for simplicity, only the stiffness is considered to be affected by the environmental conditions.

\begin{figure}[]
    \centering
        \begin{tikzpicture}
        \draw[line width=0.5mm] (-2,1) -- (-2,-1);

        \draw[line width=0.5mm] (-2, 1) -- (-2.2, 0.8);
        \draw[line width=0.5mm] (-2, 0.8) -- (-2.2, 0.6);
        \draw[line width=0.5mm] (-2, 0.6) -- (-2.2, 0.4);
        \draw[line width=0.5mm] (-2, 0.4) -- (-2.2, 0.2);
        \draw[line width=0.5mm] (-2, 0.2) -- (-2.2, 0.0);
        \draw[line width=0.5mm] (-2, 0.0) -- (-2.2, -.2);
        \draw[line width=0.5mm] (-2, -.2) -- (-2.2, -.4);
        \draw[line width=0.5mm] (-2, -.4) -- (-2.2, -.6);
        \draw[line width=0.5mm] (-2, -.6) -- (-2.2, -.8);
        \draw[line width=0.5mm] (-2, -.8) -- (-2.2, -1.0);
    
        \node (1) at (0.0, 0.0) [draw, line width=0.5mm, minimum width=1cm, minimum height=1cm] {$m$};
        \node (2) at (2.5, 0) [draw, line width=0.5mm, minimum width=1cm, minimum height=1cm] {$m$};
        \node (3) at (5.0, 0.0) [draw, line width=0.5mm, minimum width=1cm, minimum height=1cm] {$m$};
        \node (4) at (7.5, 0.0) [draw, line width=0.5mm, minimum width=1cm, minimum height=1cm] {$m$};
        \node (5) at (10.0, 0.0) [draw, line width=0.5mm, minimum width=1cm, minimum height=1cm] {$m$};
        \node (6) at (12.5, 0.0) [draw, line width=0.5mm, minimum width=1cm, minimum height=1cm] {$m$};
        
        \draw[thick, decoration={aspect=0.65, segment length=3mm,
             amplitude=0.2cm, coil}, decorate] (-2, 0) -- (-0.5, 0);
        
        \draw[thick, decoration={aspect=0.65, segment length=3mm,
             amplitude=0.2cm, coil}, decorate] (1) --(2);
        \draw[thick, decoration={aspect=0.65, segment length=3mm,
             amplitude=0.2cm, coil}, decorate] (2) --(3);
        \draw[thick, decoration={aspect=0.65, segment length=3mm,
             amplitude=0.2cm, coil}, decorate] (3) --(4);
        \draw[thick, decoration={aspect=0.65, segment length=3mm,
             amplitude=0.2cm, coil}, decorate] (4) --(5);
        \draw[thick, decoration={aspect=0.65, segment length=3mm,
             amplitude=0.2cm, coil}, decorate] (5) -- (6);

        \node[] at (-1.25, 1.0) {$k(\bm{e})$};
        \node[] at (1.25, 1.0) {$k(\bm{e})$};
        \node[] at (3.75, 1.0) {$k(\bm{e})$};
        \node[] at (6.25, 1.0) {$k(\bm{e})$};
        \node[] at (8.75, 1.0) {$k(\bm{e})$};
        \node[] at (11.25, 1.0) {$k(\bm{e})$};
        
        \draw[line width=0.5mm] (0, -0.5) -- (0, -1.0) -- (-0.5, -1.0);
        \draw[line width=0.5mm] (-0.5, -0.9) -- (-0.5, -1.1);
        \draw[line width=0.5mm]  (-0.4, -1.2) -- (-0.6, -1.2) -- (-0.6, -0.8) --                             (-0.4, -0.8);
        \draw[line width=0.5mm] (-0.6, -1.0) -- (-0.9, -1.0) -- (-0.9, -1.6);
        \draw[line width=0.5mm] (-1.1, -1.6) -- (-.7, -1.6);
        
        \draw[line width=0.5mm] (2.5, -0.5) -- (2.5, -1.0) -- (2.0, -1.0);
        \draw[line width=0.5mm] (2.0, -0.9) -- (2.0, -1.1);
        \draw[line width=0.5mm]  (2.1, -1.2) -- (1.9, -1.2) -- (1.9, -0.8) --                             (2.1, -0.8);
        \draw[line width=0.5mm] (1.9, -1.0) -- (1.6, -1.0) -- (1.6, -1.6);
        \draw[line width=0.5mm] (1.4, -1.6) -- (1.8, -1.6);
        
        \draw[line width=0.5mm] (5, -0.5) -- (5, -1.0) -- (4.5, -1.0);
        \draw[line width=0.5mm] (4.5, -0.9) -- (4.5, -1.1);
        \draw[line width=0.5mm]  (4.6, -1.2) -- (4.4, -1.2) -- (4.4, -0.8) --                             (4.6, -0.8);
        \draw[line width=0.5mm] (4.4, -1.0) -- (4.1, -1.0) -- (4.1, -1.6);
        \draw[line width=0.5mm] (3.9, -1.6) -- (4.3, -1.6);
        
        \draw[line width=0.5mm] (7.5, -0.5) -- (7.5, -1.0) -- (7.0, -1.0);
        \draw[line width=0.5mm] (7.0, -0.9) -- (7.0, -1.1);
        \draw[line width=0.5mm]  (7.1, -1.2) -- (6.9, -1.2) -- (6.9, -0.8) --                             (7.1, -0.8);
        \draw[line width=0.5mm] (6.9, -1.0) -- (6.6, -1.0) -- (6.6, -1.6);
        \draw[line width=0.5mm] (6.4, -1.6) -- (6.8, -1.6);
        
        \draw[line width=0.5mm] (10, -0.5) -- (10, -1.0) -- (9.5, -1.0);
        \draw[line width=0.5mm] (9.5, -0.9) -- (9.5, -1.1);
        \draw[line width=0.5mm]  (9.6, -1.2) -- (9.4, -1.2) -- (9.4, -0.8) --                             (9.6, -0.8);
        \draw[line width=0.5mm] (9.4, -1.0) -- (9.1, -1.0) -- (9.1, -1.6);
        \draw[line width=0.5mm] (8.9, -1.6) -- (9.3, -1.6);
        
        \draw[line width=0.5mm] (12.5, -0.5) -- (12.5, -1.0) -- (12.0, -1.0);
        \draw[line width=0.5mm] (12.0, -0.9) -- (12.0, -1.1);
        \draw[line width=0.5mm]  (12.1, -1.2) -- (11.9, -1.2) -- (11.9, -0.8) --                             (12.1, -0.8);
        \draw[line width=0.5mm] (11.9, -1.0) -- (11.6, -1.0) -- (11.6, -1.6);
        \draw[line width=0.5mm] (11.4, -1.6) -- (11.8, -1.6);
        
        \node[] at (-1.25, -1.0) {$c$};
        \node[] at (1.25, -1.0) {$c$};
        \node[] at (3.75, -1.0) {$c$};
        \node[] at (6.25, -1.0) {$c$};
        \node[] at (8.75, -1.0) {$c$};
        \node[] at (11.25, -1.0) {$c$};

    \end{tikzpicture}
    \caption{Example of a mass-spring system, with masses $m$, damping coefficients $c$ and spring stiffness $k$, which is a function of the environmental conditions $\bm{e}$.}
    \label{fig:mass_spring_example}
\end{figure}
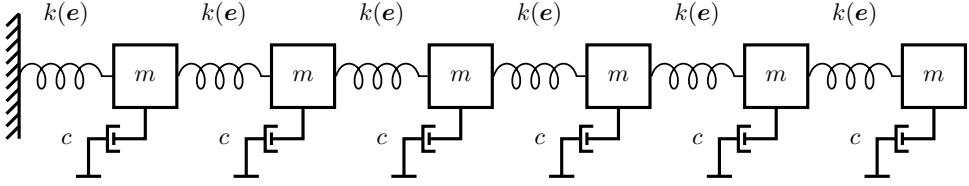

\begin{figure}
    \centering
    \begin{tikzpicture}
        \definecolor{magenta1}{RGB}{204, 0, 153}
        \definecolor{green1}{RGB}{51, 153, 51}
        \definecolor{orange1}{RGB}{255, 153, 51}

        \draw[-{>[scale=2.5, length=2, width=3]}, line width=0.5mm] (0.0, 1.0) to (0.0, 4.0);
        \draw[-{>[scale=2.5, length=2, width=3]}, line width=0.5mm] (-0.1, 1.1) to (5.5, 1.1);
        
        \node[] (l) at (-0.6, 3.6) {$\pazocal{L}$};
        
        \node[] (param) at (5.5, 0.7) {$\bm{\theta}$};

        \draw[line width=0.3mm, magenta1] (0.0, 2.8) to[out=-10,in=180] (1.0, 1.5) to[out=0,in=210] (2.0, 3.0) to[out=30,in=180] (3.5, 2.0) to[out=0,in=220] (5.0, 3.8);
        
        \fill[green1] (3.5, 2.0) circle (0.15cm);
        
        \fill[orange1] (1.0, 1.5) circle (0.15cm);

    \end{tikzpicture}
    \caption{Schematic representation of an objective function $\pazocal{L}$ as a function of the trainable parameter $\theta$ and two local minima shown in orange and green.}
    \label{fig:loss}
\end{figure}
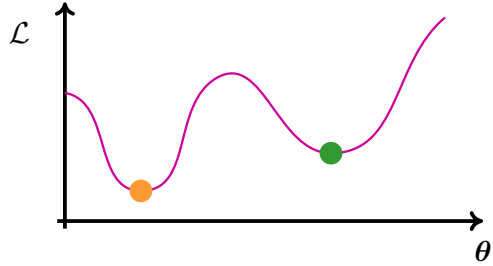

For a single structure, a neural network, can be trained to predict some features of the system; e.g. its natural frequencies, a \textit{frequency response function} (FRF), a \textit{transmissibility} etc. This is a single task $\tau_{i}$ and can be effectively dealt with (and has been performed in the past), given enough training data denoted as,
\begin{equation}
    \pazocal{D}_{i} = \{(\bm{e}_{j}, \bm{y}_{j})\}, \quad i=1, 2... n, \quad j = 1, 2... m
\end{equation}
where $\bm{e}_{j}$ is the $j$\textsuperscript{th} instance of environmental conditions and $\bm{y}_{j}$ the corresponding vector of quantities that are being modelled. To illustrate the desired form of data-driven models, in Figure \ref{fig:loss}, the value of the loss function $\pazocal{L}$ is shown (the magenta curve) as a function of the trainable parameters $\bm{\theta}$ (the parameter space will actually be multidimensional with very high dimensionality for neural networks, but for the sake of simplicity and visualisation it is depicted as one-dimensional). In the same figure, two local minima are also depicted; whether they are local or global minima is not of interest for the current case study, as long as the accuracy of the model for the specific set of parameters is acceptable. Consider the values of the loss function, as well as of the minima, to correspond to the whole range of input values: i.e.\ the value of the loss $\pazocal{L}$ is the \textit{total discrepancy} between all the potential predictions of the model and the true underlying relationship of the environmental conditions $\bm{e}$ and the quantity $\bm{y}$ predicted.

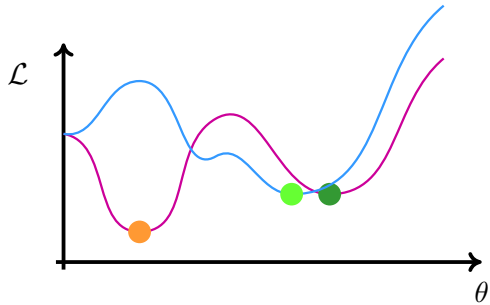
\begin{figure}
    \centering
    \begin{tikzpicture}
        \definecolor{magenta1}{RGB}{204, 0, 153}
        \definecolor{green1}{RGB}{51, 153, 51}
        \definecolor{green2}{RGB}{102, 255, 51}
        \definecolor{orange1}{RGB}{255, 153, 51}
        \definecolor{blue1}{RGB}{51, 153, 255}

        \draw[-{>[scale=2.5, length=2, width=3]}, line width=0.5mm] (0.0, 1.0) to (0.0, 4.0);
        \draw[-{>[scale=2.5, length=2, width=3]}, line width=0.5mm] (-0.1, 1.1) to (5.5, 1.1);
        
        \node[] (l) at (-0.6, 3.6) {$\pazocal{L}$};
        
        \node[] (param) at (5.5, 0.7) {$\theta$};

        \draw[line width=0.3mm, magenta1] (0.0, 2.8) to[out=-10,in=180] (1.0, 1.5) to[out=0,in=210] (2.0, 3.0) to[out=30,in=180] (3.5, 2.0) to[out=0,in=220] (5.0, 3.8);
        
        \fill[green1] (3.5, 2.0) circle (0.15cm);
        
        \fill[orange1] (1.0, 1.5) circle (0.15cm);
        
        \draw[line width=0.3mm, blue1] (0.0, 2.8) to[out=-10,in=180] (1.0, 3.5) to[out=0,in=210] (2.0, 2.5) to[out=30,in=180] (3.0, 2.0) to[out=0,in=220] (5.0, 4.5);
        
        \fill[green2] (3.0, 2.0) circle (0.15cm);
        
    \end{tikzpicture}
    \caption{Objective function $\pazocal{L}$ for two tasks (magenta and blue lines) as a function of the trainable parameter $\Theta$ and three local minima shown in orange, green and light green.}
    \label{fig:loss_2}
\end{figure}

The orange point corresponds to a lower loss value, meaning that the corresponding parameter values are the optimal choice for the current task. The green point is also a local minimum and its accuracy may also be acceptable for the purposes of the task. Considering the system within a population-based scheme, another structure is introduced, which is to be modelled using the same model (i.e. the same architecture neural network), and the same loss function. The newly-introduced loss function landscape is shown in Figure \ref{fig:loss_2} (blue curve). As shown, the parameters which corresponded to the minimum for the previous structure, do not correspond to a local minimum. However, one of the two minima may have moved to a neighbouring point, the light green point in the same figure. 

This example is case specific, but aims at demonstrating a problem that may often occur when training data-driven models in a population-based scheme. When considering a new task/structure, the minima of the first task most probably will not be minima for the new task. However, some minima may transition smoothly to neighbouring points of the parameter space, as the underlying parameters that describe the tasks vary. Therefore, a family of models may describe a variety of tasks, with their trainable parameters $\bm{\theta}$ lying on a specific area/manifold of the parameter space. As the tasks vary, so shall the parameters that correspond to the local minimum, but always on the same manifold/area of the parameter space.

\begin{figure}
    \centering
    \begin{tikzpicture}[scale=0.9, every node/.style={scale=0.9}]
     \definecolor{blue1}{RGB}{93, 143, 218}
     \definecolor{teal}{RGB}{100, 225, 225}
     \definecolor{gray1}{RGB}{127, 152, 154}
     \definecolor{green1}{RGB}{76, 153, 0}
     \definecolor{orange1}{RGB}{255, 153, 51}
     \definecolor{magenta1}{RGB}{204, 0, 153}
     % Base manifold M
     \draw[line width=0.2mm, black] (0.0, 0.0) to[out=30, in=150] (3.0, 0.5) to[out=330, in=210] (6.0, 1.0);
     \draw[line width=0.2mm, black] (6.0, 1.0) to[] (7.0, 3.0);
     \draw[line width=0.2mm, black] (1.0, 2.0) to[out=30, in=150] (3.0, 2.5) to[out=330, in=210] (7.0, 3.0);
     \draw[line width=0.2mm, black] (1.0, 2.0) to[] (0.0, 0.0);

    % Fibre 1
      \draw[dashed, line width=0.15mm, magenta1] (1.2, 1.0) to (1.2, 4.8);
      \draw[line width=0.15mm, magenta1] (1.2, 4.8) to (1.2, 9.2);

      \fill[magenta1] (1.2, 4.8) circle (0.08cm);
      \fill[magenta1] (1.2, 9.2) circle (0.08cm);

      \fill[orange1] (1.2, 7.6) circle (0.08cm);

      \fill[green1] (1.2, 6.1) circle (0.08cm);

    % Fibre 2
      \draw[dashed, line width=0.15mm, magenta1] (4.4, 1.4) to (4.4, 5.2);
      \draw[line width=0.15mm, magenta1] (4.4, 5.2) to (4.4, 10.2);

      \fill[magenta1] (4.4, 5.4) circle (0.08cm);
      \fill[magenta1] (4.4, 10.2) circle (0.08cm);

      \fill[orange1] (4.4, 8.8) circle (0.08cm);

      \fill[green1] (4.4, 6.15) circle (0.08cm);

    % Fibre 3
      \draw[dashed, line width=0.15mm, magenta1] (2.9, 1.1) to (2.9, 5.0);
      \draw[line width=0.15mm, magenta1] (2.9, 5.0) to (2.9, 9.7);

      \fill[magenta1] (2.9, 5.0) circle (0.08cm);
      \fill[magenta1] (2.9, 9.7) circle (0.08cm);

      \fill[orange1] (2.9, 7.35) circle (0.08cm);

      \fill[green1] (2.9, 6.0) circle (0.08cm);

     % Lis
     \node[] at (3.25, 9.75) {\textcolor{magenta1}{$\pazocal{L}_{2}$}};

     \node[] at (4.75, 10.25) {\textcolor{magenta1}{$\pazocal{L}_{i}$}};

     \node[] at (1.55, 9.25) {\textcolor{magenta1}{$\pazocal{L}_{1}$}};

      % Tis
     \node[] at (3.25, 1.15) {\textcolor{black}{$\tau_{2}$}};

     \node[] at (4.75, 1.45) {\textcolor{black}{$\tau_{i}$}};

     \node[] at (1.55, 1.05) {\textcolor{black}{$\tau_{1}$}};

     \fill[black] (2.9, 1.1) circle (0.08cm);
     \fill[black] (4.4, 1.4) circle (0.08cm);
     \fill[black] (1.2, 1.0) circle (0.08cm);
     
     % Total space E
     
     \draw[line width=0.2mm, blue1] (0.0, 8.0) to[out=60, in=150] (2.0, 8.75) to[out=330, in=200] (4.0, 9.25) to[out=20, in=130] (6.0, 8.5);
     \draw[line width=0.2mm, blue1] (0.0, 4.0) to[] (0.0, 8.0);
     \draw[line width=0.2mm, blue1] (6.0, 4.5) to[] (6.0, 8.5);
     
     \draw[line width=0.2mm, blue1] (0.0, 8.0) to[] (1.0, 9.5);
     \draw[line width=0.2mm, blue1] (6.0, 8.5) to[] (7.0, 10.0);
     
     \draw[line width=0.2mm, blue1] (1.0, 9.5) to[out=60, in=150] (3.0, 10.25) to[out=330, in=200] (5.0, 10.75) to[out=20, in=130] (7.0, 10.0);
     
     \draw[line width=0.2mm, blue1] (6.0, 4.5) to[] (7.0, 6.0);
     \draw[line width=0.2mm, blue1] (7.0, 6.0) to[] (7.0, 10.0);
     
     \node[] (M) at (7.0, 2.0) {$\pazocal{T}$};
     
     \node[] (E) at (7.6, 7.0) {$\pazocal{L} \times \pazocal{T}$};

     % \fill[gray1, fill opacity=0.4] (0.0, 5.0) to (1.0, 6.5) to[out=60, in=150] (3.0, 6.75) to[out=330, in=240] (5.0, 6.75) to[out=60, in=130] (7.0, 7.0) to (6.0, 5.5) to[out=130, in=60] (4.0, 5.25) to[out=240, in=330] (2.0, 5.25) to[out=150, in=60] (0.0, 5.0);

     % mini areas 1
     \draw[line width=0.3mm, orange1] (0.4, 7.4) to[out=75, in=170] (0.7, 7.7) to[out=-10, in=240] (1.0, 8.0) to[out=30, in=190] (1.4, 8.2) to[out=10, in=160] (1.8, 8.0) to[out=260, in=45] (1.6, 7.7) to[out=225, in=45] (1.2, 7.4) to[out=160, in=45] (0.8, 7.3) to[out=225, in=-5] (0.4, 7.4);

    \fill[orange1, fill opacity=0.4] (0.4, 7.4) to[out=75, in=170] (0.7, 7.7) to[out=-10, in=240] (1.0, 8.0) to[out=30, in=190] (1.4, 8.2) to[out=10, in=160] (1.8, 8.0) to[out=260, in=45] (1.6, 7.7) to[out=225, in=45] (1.2, 7.4) to[out=160, in=45] (0.8, 7.3) to[out=225, in=-5] (0.4, 7.4);

      % mini areas 2
      \draw[line width=0.3mm, orange1] (3.4, 8.4) to[out=-20, in=230] (3.7, 8.6) to[out=40, in=180] (4.0, 9.0) to[out=-30, in=200] (4.4, 9.0) to[out=20, in=160] (4.8, 9.0) to[out=260, in=45] (4.6, 8.8) to[out=225, in=60] (4.2, 8.4) to[out=160, in=45] (3.8, 8.3) to[out=225, in=-30] (3.4, 8.4);

      \fill[orange1, fill opacity=0.4] (3.4, 8.4) to[out=-20, in=230] (3.7, 8.6) to[out=40, in=180] (4.0, 9.0) to[out=-30, in=200] (4.4, 9.0) to[out=20, in=160] (4.8, 9.0) to[out=260, in=45] (4.6, 8.8) to[out=225, in=60] (4.2, 8.4) to[out=160, in=45] (3.8, 8.3) to[out=225, in=-30] (3.4, 8.4);

      % mini areas 3
      \draw[line width=0.3mm, orange1] (2.4, 7.2) to[out=-20, in=230] (2.7, 7.4) to[out=40, in=180] (3.0, 7.8) to[out=-30, in=200] (3.4, 7.8) to[out=20, in=140] (3.8, 7.8) to[out=260, in=45] (3.6, 7.6) to[out=225, in=60] (3.2, 6.8) to[out=160, in=45] (2.8, 7.1) to[out=225, in=-30] (2.4, 7.2);

      \fill[orange1, fill opacity=0.4] (2.4, 7.2) to[out=-20, in=230] (2.7, 7.4) to[out=40, in=180] (3.0, 7.8) to[out=-30, in=200] (3.4, 7.8) to[out=20, in=140] (3.8, 7.8) to[out=260, in=45] (3.6, 7.6) to[out=225, in=60] (3.2, 6.8) to[out=160, in=45] (2.8, 7.1) to[out=225, in=-30] (2.4, 7.2);

     \draw[line width=0.2mm, blue1] (0.0, 4.0) to[out=60, in=150] (2.0, 4.25) to[out=330, in=240] (4.0, 4.25) to[out=60, in=130] (6.0, 4.5);
     
     % proper big area
     \draw[line width=0.3mm, green1] (0.0, 5.0) to (1.0, 6.5) to[out=60, in=150] (3.0, 6.75) to[out=330, in=240] (5.0, 6.75) to[out=60, in=130] (7.0, 7.0)
     to[out=190, in=60](6.3, 6.8) to[out=240, in=45] (6.0, 5.5) to[out=160, in=30] (4.0, 5.55) to[out=210, in=330] (2.0, 5.65) to[out=150, in=60] (0.0, 5.0);

     \fill[green1,  fill opacity=0.4] (0.0, 5.0) to (1.0, 6.5) to[out=60, in=150] (3.0, 6.75) to[out=330, in=240] (5.0, 6.75) to[out=60, in=130] (7.0, 7.0)
     to[out=190, in=60](6.3, 6.8) to[out=240, in=45] (6.0, 5.5) to[out=160, in=30] (4.0, 5.55) to[out=210, in=330] (2.0, 5.65) to[out=150, in=60] (0.0, 5.0);

\end{tikzpicture}
    \caption{Schematic of a fibre bundle of the loss functions of a data-driven model across a population of tasks. The base manifold $\pazocal{T}$ is formed by points representing the tasks $\tau_{i}$ of the population. To every task $\tau_{i}$, a fibre $\pazocal{L}_{i}$ corresponds, which represents the values of the loss function $\pazocal_{L}$ for different values of the trainable parameters of the data-driven model. The orange and green points represent parts of the fibre bundle where local or global minima are observed regarding the loss function. The fibres intercepting an orange area have a local or global minimum at the intersection point between the fibre and the orange area. The same applies for the green area. However, the green area is a continuous cross section along the whole bundle.}
    \label{fig:loss_bundle}
\end{figure}
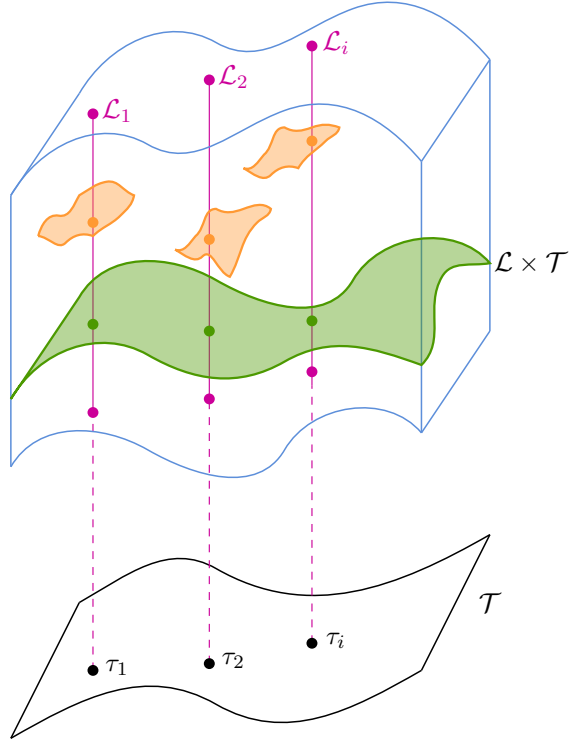

To further visualise the idea above, a \textit{fibre bundle} \cite{Schutz, Hamilton} is considered. The base manifold of the fibre bundle is the manifold of the tasks $\pazocal{T}$. On this manifold various tasks $\tau_{i}$ are located. The fibre corresponding to each point $\tau_{i}$ of the base manifold, is the manifold shaped by the loss function $\pazocal{L}$ on the parameter space $\Theta$. Consider the set of points of the fibre bundle, which is the collection of local minima for each task. Among these points, the desired models lie on a \textit{cross section}, i.e. a collection of points, across the whole bundle, which are glued together. Given such a cross section, when one needs to move to a neighbouring task, a local minimum for the new task shall exist within the neighbourhood of the current task. The idea is depicted in Figure \ref{fig:loss_bundle}. In the figure, some tasks $\tau_{i}$ are shown with their corresponding fibres (magenta vertical lines), defined as the corresponding losses in the parameter space. The green- and orange-coloured points represent local minima on these functions. As shown, for the orange local minima, as one moves towards neighbouring tasks, the minimum remains glued to the minimum for a small area of the task space, but after some point, it is not a minimum anymore. In contrast, for the green minima, one can navigate through the whole manifold $\pazocal{T}$ and for each step in the neighbour of the current task, a local minimum will also exist in the neighbourhood of the current local minimum.

Fibre bundle shaped using as base manifold the manifold $\pazocal{T}$ of tasks $\tau_{i}$, each fibre corresponds to the loss function manifold $\pazocal{L}_{i}$ in the parameter space $\theta$. Also, local minima are shown with coloured points (green and orange).

From a physical point of view, the idea above coincides with the idea of forcing the model to approximate the underlying physics of the population of systems. A neural network may be trained to have acceptable accuracy for the task in hand, but this does not mean that it has learnt the underlying physics of the problem. Such generalisation problems may occur when the training data come from specific operational conditions but the testing data come from different operational conditions. According to the idea presented, one can consider that the points on the green cross-section in Figure \ref{fig:loss_bundle} correspond to a family of models that have learnt the underlying physics of the population. The orange points cannot be considered appropriate captures of the underlying physics, because as the parameters of the systems change smoothly, the values of the trainable parameters of the model, which correspond to a local minimum, should also change smoothly; however, these points do not have this property.

\begin{figure}
    \centering
    \includegraphics[scale=0.45]{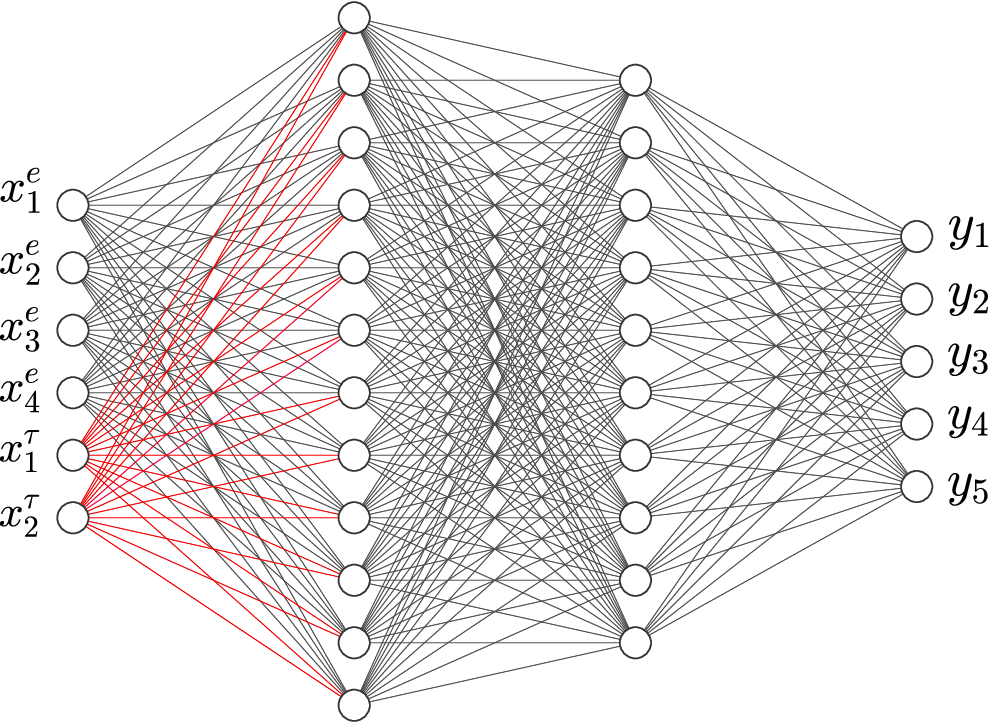}
    \caption{Neural network modelling the population with inputs regarding the external variables $\bm{x}^{e}$ and the structure $\bm{x}^{\tau}$.}
    \label{fig:nn_with_tau}
\end{figure}

\begin{figure}
    \centering
    \includegraphics[scale=0.45]{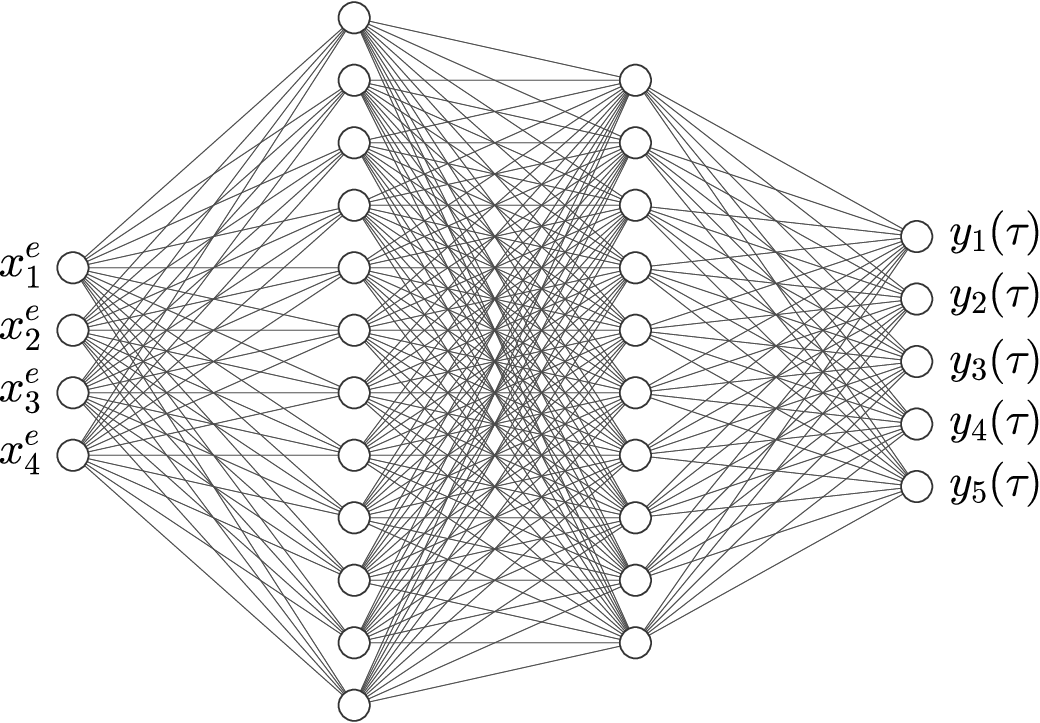}
    \caption{Neural network modelling the population with inputs regarding only the external variables $\bm{x}^{e}$.}
    \label{fig:nn_no_tau}
\end{figure}

Although the existence of such a cross section in the fibre bundle is not guaranteed, it might exist for a population of structures with common underlying physics. Consider the case of building a neural network model which takes into account the parametrisation of the structures and aims at approximating the behaviour of the structures of the population - similar to the model $f^{d}$ of equation (\ref{eq:data_driven_2}). In that case, $f^{d}$ approximates some quantities of interest $\bm{y}$ as,
\begin{equation}
    \bm{y} = f^{d}(\bm{x}^{e}, \bm{x}^{\tau})
\end{equation}
where $\bm{x}^{e}$ are the inputs describing the external/environmental parameters and $\bm{x}^{\tau}$ are the input variables describing the structure/task. Let $f^{d}$ be a feedforward neural network with specified architecture and, without loss of generality, consider it to be the neural network of Figure \ref{fig:nn_with_tau}. Consider a second family of neural networks $\bar{f}^{d}$ of the same architecture of hidden layers and output layer but with input variables only the environmental variables $\bm{x}^{e}$, defined as,
\begin{equation}
    \bm{y}(\tau) = \bar{f}^{d}(\bm{x}^{e})
\end{equation}
where the output is a function of the task $\tau$. Without loss of generality, let the $\bar{f}^{d}$ neural networks be like the one shown in Figure \ref{fig:nn_no_tau}. The neural networks $f^{d}$ and $\bar{f}^{d}$ can be equivalent for each task $\tau$ and the corresponding inputs $\bm{x}^{\tau}$ by setting their common trainable parameters to be equal, except for a small subset of them. The common trainable parameters of the two neural networks of Figures \ref{fig:nn_with_tau} and \ref{fig:nn_no_tau} are the weights of the grey connections and the biases of their neurons. For some task $\tau_{i}$, the input values $\bm{x}^{\tau}$ for the first neural network are constant. Therefore, in the second neural network, for the same task $\tau_{i}$, by varying some of its trainable parameters to account for the missing contribution of $\bm{x}^{\tau}$, one can get a neural network equivalent to $f^{d}$ for constant $\bm{x}^{\tau}$ and varying $\bm{x}^{e}$; a trivial solution is to vary the bias terms of the hidden layer to compensate for the effect of the missing neural-network connections of $\bm{x}^{\tau}$, the red connections in Figure \ref{fig:nn_with_tau}. Denoting the trainable parameters of $\bar{f}^{d}$, which need to vary to capture the effects of varying $\bm{x}^{\tau}$, as $\bm{\theta_{v}}$ yields,
\begin{equation}
    f^{d}(\bm{x}^{e}, \bm{x}^{\tau}) = \bar{f}^{d}(\bm{x}^{e}, \bm{\theta_{v}})
\end{equation}
Computing the derivatives with respect to $\bm{x}^{\tau}$, and taking into account the chain rule on the right hand side, yields,
\begin{equation}
    \frac{\partial f^{d}}{\partial \bm{x}^{\tau}} = \frac{\partial \bar{f}^{d}}{\partial \bm{\theta}_{v}
    } \frac{\partial \bm{\theta}_{v}}{\partial \bm{x}^{\tau}}
\end{equation}
The quantities $\frac{\partial f^{d}}{\partial \bm{x}^{\tau}}$ and $\frac{\partial \bar{f}^{d}}{\partial \bm{\theta}_{v}}$ are continuous for continuous activation functions (or piece-wise continuous for activation functions such as relu or leaky-relu), since they are derivatives of the output of neural networks with respect to some of the input variables or some of their trainable parameters. As a result $\frac{\partial \bm{\theta}_{v}}{\partial \bm{x}^{\tau}}$ should also be continuous, except for the case that $\frac{\partial \bar{f}^{d}}{\partial \bm{\theta}_{v}}$ is equal to zero. However, the derivative of the output of a neural network with respect to its trainable parameters is rarely exactly equal to zero, given that the parameters $\bm{\theta}_{v}$ emulate the effect of the task-describing variables $\bm{x}^{\tau}$ they are not expected to be zero, since the task affects largely the quantities $\bm{y}$. As a result $\frac{\partial \bm{\theta}_{v}}{\partial \bm{x}^{\tau}}$ is indeed continuous, meaning that, as the parameters $\bm{x}^{\tau}$ that characterise the task $\tau$ vary smoothly, so do the trainable parameters $\bm{\theta}_{v}$ of the $\bar{f}^{d}$ neural network, for which $\bar{f}^{d}$ is equivalent to the $f^{d}$ neural network, which approximates adequately the underlying physics of the population. Although the above study is not a proof, it indicates that, if a latent set of continuous variables which describe the structures exists, the model $f^{d}$ can accommodate the changes in the behaviour by smoothly altering the values of the trainable parameters. Thus, the cross section of the fibre bundle of Figure \ref{fig:loss_bundle}, which connects the values of the trainable parameters for which error minima are achieved, may exist for a population of heterogeneous structures.

\subsection{Model-agnostic meta-learning}
\label{sec:MAML}

For reasons covered in previous sections, data-driven models often lack the ability to generalise well; this is often caused by the lack of sufficient data for proper training or because one attempts to use such models on data outside their training domain. To deal with such issues, several techniques have been used. Such techniques are often referred to as \textit{few-shot learning} \cite{wang2019few}. Methods belonging to the few-shot learning domain are: the Siamese neural networks \cite{koch2015siamese}, the triplet network \cite{hoffer2015deep}, the matching network \cite{vinyals2016matching} etc. Another way of dealing with such lack-of-data problems is \textit{transfer learning}, which attempts to solve these problems by exploiting knowledge about models trained on some domain by transferring it to another domain. This knowledge is usually in the form of some feature extractor \cite{oquab2014learning}, or by seeking a domain, onto which data from different tasks can be mapped and a common model can be used \cite{gardner2022application, chen2022wheel}.

Another approach that aims at dealing with the problem of lack of data is \textit{meta-learning} \cite{hospedales2020meta}. Meta-learning can be considered as the domain of machine learning that aims at developing algorithms that ‘learn to learn’. A general idea behind such an approach to machine learning, is that the optimisation techniques that are being used - such as gradient descent \cite{Bishop, Bishop2} or Adam \cite{kingma2014adam} - are general purpose algorithms and therefore, not targeted to the problem that one seeks to solve. Furthermore, in the case of population-based modelling, such optimisation algorithms do not exploit knowledge across the population to enhance their performance. Consequently, the goal of meta-learning is to force the training procedure to be effective for a family of tasks; a family of tasks can be a collection of similar problems, domains, or, as in the current work, a population of phenomena with similar underlying physics. A physics-based version of this procedure is described in \cite{bull2022knowledge}, where the underlying physics of the modelled phenomenon are defined via parametric functions. In the aforementioned work, the members of the population behave according to the pre-defined functions, but also according to tunable parameters, which are tuned in a Bayesian manner when data are acquired from a structure. The definition of the underlying relationship is done by the analyser, reducing the need for data to fit a model for a new structure with a few available data samples. In the current work, the proposed methods are expected to learn an underlying relationship from a population of structures via meta-learning.

A formalisation of meta-learning is given in \cite{titsias2021information}, where its objective function is defined as the maximisation of,
\begin{equation}
    \label{eq:meta_learning_formal}
    \pazocal{L} = \pazocal{I}(\pazocal{T}, D^{v}) - \beta \pazocal{I}(\pazocal{T}, D^{t})
\end{equation}
where $\pazocal{T}$ represents the different tasks (denoted as $Z$ in the original work), $D^{v}$ and $D^{t}$ represent the unseen data and the training data correspondingly, $\beta$ is a regularisation parameter and $\pazocal{I}$ is the mutual information. Maximisation of the first term of the specific objective function means maximisation of the information provided by the task $\tau_{i}$ regarding the relationship of the unseen data $D^{v}$, i.e. the model should be able to make predictions for the testing data more efficiently when the task is given or inferred. The second term, as stated in \cite{titsias2021information} is a regularisation term which aims at reducing the mutual information between the training data and the task in order to extract more informative features for the unseen data. The regularisation term aims at discouraging the model from overfitting to the training data and to learn task-independent features. Optimising a population-informed model using the above equation 

Such a scheme matches the framework proposed in the current work. In the case of having a physics-based model which describes appropriately the physics of the population, as in \cite{bull2022knowledge}, a small number of observations from a new structure should suffice to define the parameters of the model and to provide accurate predictions. Similarly, in the current work, a family of neural network models is sought, which matches the underlying physics of a population and, with only a small number of available training data, shall be able to make accurate predictions. Therefore, considering that the current task $\tau_{i}$ is described by the available training data, the maximisation of the mutual information between the available data and the unseen data, could provide such a family of neural-network models.

Meta-learning shows a similar approach to the problem of lack of data as \textit{multi-task learning} \cite{ruder2017overview}. Methods of the multi-task learning domain involve parameter sharing between the initial layers of neural networks. Examples of parameter sharing be can found in \cite{caruana1993multitask} and \cite{duong2015low}, where hard and soft parameter sharing techniques are presented respectively. Meta-learning algorithms could be considered to be a subset of multi-task learning. The main difference between the two is that meta-learning aims at developing algorithms which are able to learn with a few data samples, in contrast to multi-task learning, where often the size of the trainable parameters of a model are reduced by using heuristics, such as whole neural-network-layer transferring. Another difference is that many meta-learning algorithms are focussed on the learning procedure itself, i.e.\ the optimisation of the model parameters.

Several methods are available to perform meta-learning as discussed in \cite{hospedales2020meta}. In the current work, the method selected to perform meta-learning within a population of structures is \textit{model-agnostic meta-learning} (MAML) \cite{finn2017model}. The method aims at training a model, which can quickly - i.e.\ with a few training steps and with a few samples - adapt to a newly-presented task $\tau$, which comes from a family of tasks $\pazocal{T}$.

In the original work \cite{finn2017model}, MAML is exploited to enhance the performance of a classifier regarding the Omniglot dataset \cite{lake2011one}. The dataset comprises 1623 handwritten characters from 50 different alphabets, written by 20 different people. The resulting dataset has many different characters, but not enough to train a convolutional neural network classifier to perform sufficiently in recognising from which alphabet the characters come. 

Altering the traditional machine learning framework and creating various tasks, referred to as $K$-shot $N$-way classification, a different approach to the problem is presented. The new goal is to define classifiers which distinguish between $N$ classes of the images having $K$ samples from each class, where $K$ is often a small number and $N$ is smaller than the number of all available classes. Considering each $K$-shot $N$-way problem of the aforementioned dataset as a different task, one should seek a way to exploit knowledge from the whole population of tasks, in order to create models that are able to accurately classify the data into $N$ classes, even if the number $K$ is extremely small; in some cases $K$ may even be unity.

The algorithm in the case of classification yields quite satisfactory results; it is also tested in a regression problem and reveals promising results. In \cite{finn2017model} and \cite{cheng2022transfer} the method is used to approximate relationships which belong to a parameteric family, transferring knowledge from a set of available observations for some members of the population. In the current work, the underlying physics of the population are defined by the parameteric functions, whose exact relationship is considered to be unknown during inference. As in other works, the algorithm does not need any prior knowledge of the potential values of the parameters of the underlying relationships. 

Following \cite{finn2017model}, the algorithm here considers a model $f_{\bm{\theta}}$ with trainable parameters $\bm{\theta}$. In the original work, as well as in the current work, the model $f_{\bm{\theta}}$ is a neural network. The common way of training a neural network for a supervised-learning scheme, is to acquire a set of data $\pazocal{D}^{\tau_{i}}$ for a single task $\tau_{i}$ and perform backpropagation steps in order to find the set of parameters $\bm{\theta}^{*}$, which minimises a loss function $\pazocal{L}^{\tau_{i}}(f_{\bm{\theta}})$, summarised by,
\begin{equation}
    \bm{\theta}^{*} = \min_{\bm{\theta}}\pazocal{L}^{\tau_{i}}(f_{\bm{\theta}}; \pazocal{D}^{\tau_{i}})
\end{equation}
where the dataset $\pazocal{D}^{\tau_{i}}$ is of the form $\{(\bm{x}_{1}, \bm{y}_{1}), (\bm{x}_{2}, \bm{y}_{2})... (\bm{x}_{N}, \bm{y}_{N}) \}$ and $(\bm{x}_{j}, \bm{y}_{j})$ is the $j$\textsuperscript{th} pair of inputs $\bm{x}_{j}$ and target output values $\bm{y}_{j}$. The model with parameters $\bm{\theta}^{*}$ is considered \textit{trained} and can be used to predict values of the unseen quantities $\bm{y}^{*}$ for new values of the inputs $\bm{x}^{*}$.

% The above approach is the traditional way of applying a supervised machine learning framework, and by training several models $f^{i}$ for the various tasks $\tau_{i}$, one may get sufficient accuracy for each different task. However, it is common that for some tasks more data may be available than for other tasks, for which only a few data samples may have been acquired. In such cases, a separate-model framework would probably yield poor results for the cases with only few data available. Nevertheless, training separate models for every task is not a population-based strategy and does not exploit knowledge from the population regarding the underlying physics of the problem in order to enhance the performance of the models for all members of the population. 

The MAML approach aims at defining a model, as the initialisation point of neural networks, which shall adapt quickly to new tasks. The difference to traditional learning is that MAML encourages the model to find a point in the trainable parameter space of the neural network, from where, with a few backpropagation steps, the model should be able to converge to a task-specific set of trainable parameters quite quickly. This point is found by performing two sets of backpropagation steps. Initially backpropagation steps are performed for a set of tasks $\pazocal{T}_{t} \subseteq \pazocal{T}$. The updated parameters $\bm{\theta}'$, as a result of this first backpropagation step, the \textit{inner update}, are given by,
\begin{equation}
    \bm{\theta}' = \bm{\theta} - \alpha \nabla \pazocal{L}(f_{\bm{\theta}}; \pazocal{D}_{\tau_{i}}^{tr})
\end{equation}
where $\alpha$ is the value of the inner updates' learning rate, $\pazocal{L}$ is a loss function, $\nabla \pazocal{L} = [\frac{\partial \pazocal{L}}{\partial \theta_{1}}, \frac{\partial \pazocal{L}}{\partial \theta_{2}}, ... \frac{\partial \pazocal{L}}{\partial \theta_{i}}...]^{T}$ and $\pazocal{D}_{\tau_{i}}^{tr}$ is a dataset sampled from the task $\tau_{i}$. After performing updates for tasks $\tau_{i} \in \pazocal{T}_{t}$, the second set of updates, the \textit{meta-updates} are performed according to the loss function defined as,
\begin{equation}
    \min_{\bm{\theta}} \sum_{\tau_{i} \in \pazocal{T}} \pazocal{L}_{\tau_{i}} (f_{\bm{\theta}'})
\end{equation}
so the values of the trainable parameters $\theta$ are finally updated as,
\begin{equation}
    \bm{\theta} \gets \bm{\theta} - \beta \nabla_{\bm{\theta}} \sum_{\tau_{i} \in \pazocal{T}_{t}} \pazocal{L}(f_{\bm{\theta}'_{i}}; \pazocal{D}^{m}_{\tau_{i}})
\end{equation}
where $\beta$ is the value of the meta-updates' learning rate and $\pazocal{D}^{m}_{\tau_{i}}$ are the data sampled from task $\tau_{i}$ for the meta-update. An important aspect of the equation above is that the gradients are calculated not with respect to the updated values $\bm{\theta}_{i}$, as it would be expected, but with respect to the values of the model parameters $\bm{\theta}$ before the inner update. The procedure aims at training towards a point $\bm{\theta}^{*}$ from where task-specific model-parameter updates shall result in minimisation of the loss function. By updating the model parameters using gradients with respect to the parameters $\bm{\theta}$ instead of $\bm{\theta}'_{i}$, the error is backpropagated through the inner updates as well. As a result, information is drawn from the inner updates, making the resulting set of parameters $\bm{\theta}^{*}$ not an optimal set of parameters for all the tasks, but a set of parameters from where training updates, similar to the inner-loop updates, will result in a task-specific optimised model. The procedure is described in Algorithm \ref{alg:MAML}. Note that compared to the equations of the original work \cite{finn2017model}, the notation $p(\pazocal{T})$ is not used, since the concept of a distribution of tasks may be confusing. 

The goal of such an approach is to find an initialisation point in the parameter space for the neural network, which is ``sensitive'' to the change of task. Given such a point, one can make the model adapt quite quickly to data from a new task. At the same time, it is expected that since the model is sensitive to the change of task, it shall also be able to adapt with a few training samples, given that these training samples sufficiently characterise the task. Backpropagation derives a limited amount of information from every available data sample; therefore, following the MAML approach, one would expect to balance the lack of information, from which a small dataset $\pazocal{D}_{\tau_{t}}$ for some task $\tau_{t}$ suffers, with the information encoded from the population while training the MAML model.

The reason that such a training procedure assists in training models within a population-based scheme might be explained by the no free-lunch theorem. As described in \cite{berner2021modern}, considering a hypothesis space $\pazocal{H}$ as the total space of functions where one shall seek a solution to the a learning task, the error $\pazocal{R}(h)$ of a function $h \in \pazocal{H}$ is bounded according to,
\begin{equation}
    \pazocal{R}(h) \leq \inf_{h \in \pazocal{H}}\pazocal{R}(h) + \epsilon_{opt} + \epsilon_{stat} + \epsilon_{appr}
\end{equation}
where $\epsilon_{opt}$ is the error of the optimisation procedure and refers to the learning algorithm, $\epsilon_{stat}$ is the statistical error referring to the random selection of training data and $\epsilon_{appr}$ is the approximation error, which refers to the error regarding the subspace $\pazocal{H}_{\delta} \subseteq \pazocal{H}$ where the search is performed; for example because of the random initialisation of the trainable parameters the search might be restricted to a small subspace of the whole space of functions. The term $\inf_{f \in \pazocal{H}}\pazocal{R}(h)$ refers to the approximation capabilities of the selected model, which in the case of neural networks should be minimal, since they are universal approximators. Meta-learning algorithms such as MAML, might be a proper way of training models in a population-based framework, since they aim at reducing the total error by reducing both the approximation error and the optimisation error. The approximation error might be reduced, since the algorithm picks an initialisation point from where solutions to the tasks of the studied task family should be close. Thus the search space $\pazocal{F}_{\delta}$ is implicitly reduced, making the search easier. The optimisation error might be reduced because the outer update of the algorithm backpropagates the error through the inner optimisation step; hence, the initialisation point is not only close to a local minimum, but also only a few backpropagation steps away.

\begin{algorithm}
\caption{Model-agnostic meta-learning (MAML)}\label{alg:MAML}
\begin{algorithmic}
\Require Family of training tasks $\pazocal{T}_{tr}$, testing tasks $\pazocal{T}_{t}$ loss function $\pazocal{L}$, $\alpha$, $\beta$ learning rates, model parameters $\bm{\theta}$
\For{each training epoch}
    \For{each task $\tau_{i} \in \pazocal{T}_{tr}$}
        \State Sample a set of data $\pazocal{D}^{tr}_{\tau_{i}}$ for training of the current task $\tau_{i}$
        \State $\bm{\theta}'_{i} \gets \bm{\theta} - \alpha \nabla \pazocal{L}(f_{\bm{\theta}}; \pazocal{D}_{\tau_{i}}^{tr})$
    \EndFor
    \State Sample data for meta-training update from every task $\tau_{i} \in \pazocal{T}_{tr}$ and form the 
    \State meta-training dataset $\pazocal{D}^{m}_{\tau_{i}}$
    \State $\bm{\theta} \gets \bm{\theta} - \beta \nabla_{\bm{\theta}} \sum_{\tau_{i} \in \pazocal{T}_{t}} \pazocal{L}(f_{\bm{\theta}'_{i}}; \pazocal{D}^{m}_{\tau_{i}})$
\EndFor
\State \textbf{Testing time}
\State Sample a task of interest $\tau_{j}$ from $\pazocal{T}_{t}$ with available data $\pazocal{D}^{t}_{\tau_{j}}$
\For{each task-specific training step}
    \State $\bm{\theta} \gets \bm{\theta}^{*} - \alpha \nabla \pazocal{L}(f_{\bm{\theta}}; \pazocal{D}^{t}_{\tau_{j}})$
\EndFor
\end{algorithmic}
\end{algorithm}

\subsection{A task identification subnetwork for population-based modelling}

A way in the literature to infuse population knowledge into a learning algorithm is by using Gaussian processes and defining a proper mean and covariance matrix \cite{cross2022physics, pitchforth2021grey}. Such an approach yields quite good results, but a requirement of the method is that the mean or covariance functions can be defined by the analyser, who is imposing knowledge into the algorithm. This might not be a simple task and in some cases it might be a very difficult one; for example, in the case of multidimensional output.

Motivated by the formalisation of meta-learning, as it is given by equation (\ref{eq:meta_learning_formal}), and by the way that Gaussian processes \cite{rasmussen2003gaussian} function, an assembly of two models for population-based modelling can be considered as an attempt to define a population-specific Gaussian-process-like model. The first model shall be used to identify a vector which characterises the tasks. The second model, the main model, is created to consider the vector, which characterises the specific task, and to make predictions about the quantities of interest. The model should be able to perform this inference using a number of observed input-output variables $(\bm{x}^{e, i}, \bm{y}^{i})$. Such a consideration corresponds to the second part of the right-hand side of equation (\ref{eq:meta_learning_formal}); i.e.\ the mutual information between the available training data and each task. Although in \cite{titsias2021information}, the quantity is set to be minimised, in the current work it is believed that its maximisation would benefit the training of the population-based model. Maximisation of the aforementioned mutual information should yield a model which is able to identify the task from a few available data samples and use this information to effectively perform inference for unseen data.

Such a framework has been developed, termed \textit{conditional neural processes} (CNP) in \cite{garnelo2018conditional}. The framework aims at imitating the behaviour of a GP, which uses some training data $\pazocal{D}^{tr} = \{(\bm{x}_{1}, \bm{y}_{1}), (\bm{x}_{2}, \bm{y}_{2})..., (\bm{x}_{N}, \bm{y}_{N}) \}$, called \textit{context points} in the original work, as its base to make predictions about new input variables $\bm{x}$ of interest. Although the basic formulation of a GP is considered to be a black-box model, it could be argued that it is not. As mentioned, a GP requires some training data as well as the selection of a \textit{mean} and a \textit{kernel} function \cite{rasmussen2003gaussian}. The latter two functions can be defined by the analyser and can be parametric functions whose parameters can be learnt from the available data. However, the form of the kernel function has to be \textit{a priori} defined, e.g.\ a Matern kernel, a squared-exponential kernel, a linear kernel or a periodic one. In order to select one of these functions, one would have to impose one's intuition and prior physical knowledge into the problem. However, if no prior knowledge exists, a general purpose kernel, such as a Gaussian kernel can be used, which is appropriate for regression and whose smoothness can be inferred or predefined.

\begin{figure}
    \centering
    \includegraphics[scale=0.45]{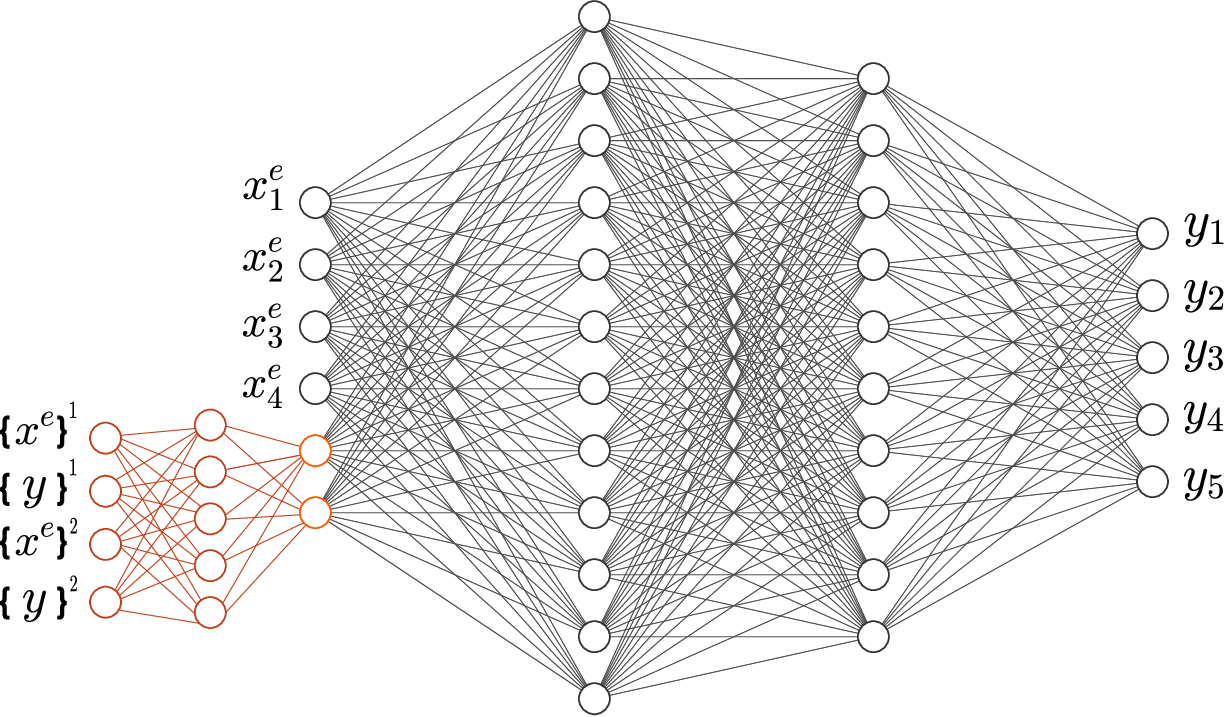}
    \caption{Neural network modelling the population with inputs regarding the external variables $\bm{x}^{e}$ and using a task identification subnetwork to map available data into some replacement variables for the task-descriptive variables.}
    \label{fig:nn_task_id}
\end{figure}

The CNP neural network can be split in two subnetworks. The main network is the one that makes predictions given the varying environmental variables $\bm{x}^{e}$ and an embedding of the task-descriptive variables $\bm{x}^{\tau}$. The embedding of the task-descriptive variable is given by the second subnetwork, which infers an arbitrary set of task-descriptive variables exploiting information from pairs of inputs and outputs of a newly-presented task. More specifically, considering available data $\pazocal{D}^{i} = \{(\bm{x}_{1}, \bm{y}_{1}), (\bm{x}_{2}, \bm{y}_{2})..., (\bm{x}_{N}, \bm{y}_{N}) \}$ for a task $\tau_{i}$, then the CNP provides predictions of the quantity of interest given by,
\begin{equation}
    \bm{y} = f^{d}(\bm{x}^{e}, \bm{x}_{1}, \bm{x}_{2}..., \bm{x}_{N}, \bm{y}_{1}, \bm{y}_{2}..., \bm{y}_{N}) = f(\bm{x}^{e}, \hat{\bm{x}}^{\tau})
\end{equation}
where $\bm{x}^{e}$ is the value of the environmental variables for which a prediction is needed, $f^{d}$ is the CNP main network and $\hat{\bm{x}}^{\tau}$ is the output of the task-identification subnetwork. An example of such an assembly is shown in Figure \ref{fig:nn_task_id}. In the example, the subnetwork shown in orange lines is the task-identification subnetwork. In the specific case, the two available pairs of known input and output vectors for the newly-presented task are, $(\bm{x}^{e, 1}, \bm{y}^{e, 1})$ and $(\bm{x}^{e, 2}, \bm{y}^{e, 2})$. Although the model in the original work is not considered a meta-learning method, one could argue that the model essentially learns to perform the training procedure of a GP and can therefore be included in the meta-learning discipline. The model performs qualitatively the functionality described by equation (\ref{eq:meta_learning_formal}), with a negative $\beta$ and deals with the problem of parametrisation of the structures, in a more direct way than MAML.

Because the models are connected in such a manner, training such a model is done simply by backpropagating the error between the output and the target vectors. For the framework of the current paper, a CNP model can be trained by randomly sampling inputs to the task-identification subnetwork, input variables $\bm{x}^{e}$ and target values $\bm{y}$ from the data-rich structures. Then the models can be tested on structures for which only a few points are available and make predictions for new input variables. In contrast to MAML, the training of a CNP is done in a traditional neural-network-training manner. The dataset is formed by randomly sampling points from the available dataset and backpropagation is repeated for several epochs.

\section{Applications}
\label{sec:applications}

In order to evaluate the efficiency of the two discussed algorithms in creating models that learn based on the population physics, an application on simulated data is presented. The application refers to a population of simulated lumped-mass systems similar to the ones shown in Figure \ref{fig:mass_spring_example}. Each structure of the population has a stiffness parameter $k$ uniformly sampled from the interval $[8000, 12000]$. The structures of the population are excited at their first degree of freedom by a white noise signal and the simulations were performed using fourth-order Runge-Kutta integration. The temperature is taken into account as an environmental parameter affecting the structures. The way that temperature is affecting the structures is by increasing or decreasing the value of the stiffness parameter of the first three springs of the system. This is considered to be an imitation of the fact that structures are often heated disproportionally throughout their volume, for example a bridge is often heated more on its deck than its pillars because of the sun radiating on the top of the bridge. 

\begin{figure}[H]
    \centering
    \includegraphics[scale=0.50]{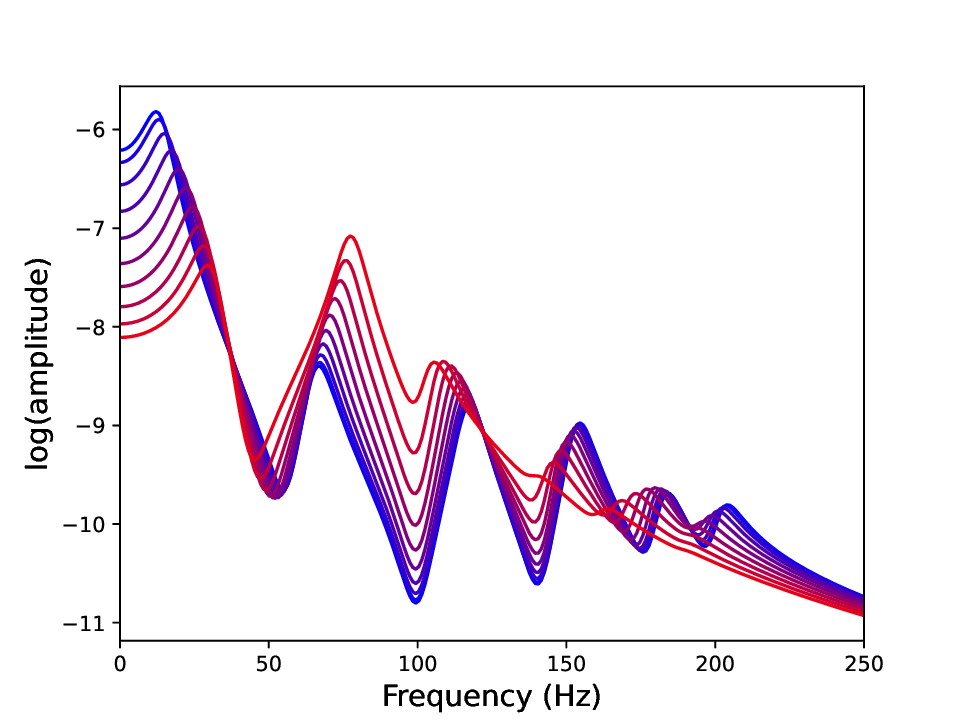}
    \caption{Frequency response functions of the first degree of freedom of a structure of the population for varying temperatures. Lower temperatures correspond to blue curves and as the temperature rises, the transition of the FRF is shown as a gradual change to red colour. Note that the changes are quite substantial because the purpose of the application is to illustrate the potential of the algorithms.}
    \label{fig:varying_temp_FRFs}
\end{figure}

\begin{figure}[H]
    \centering
    \includegraphics[scale=0.40]{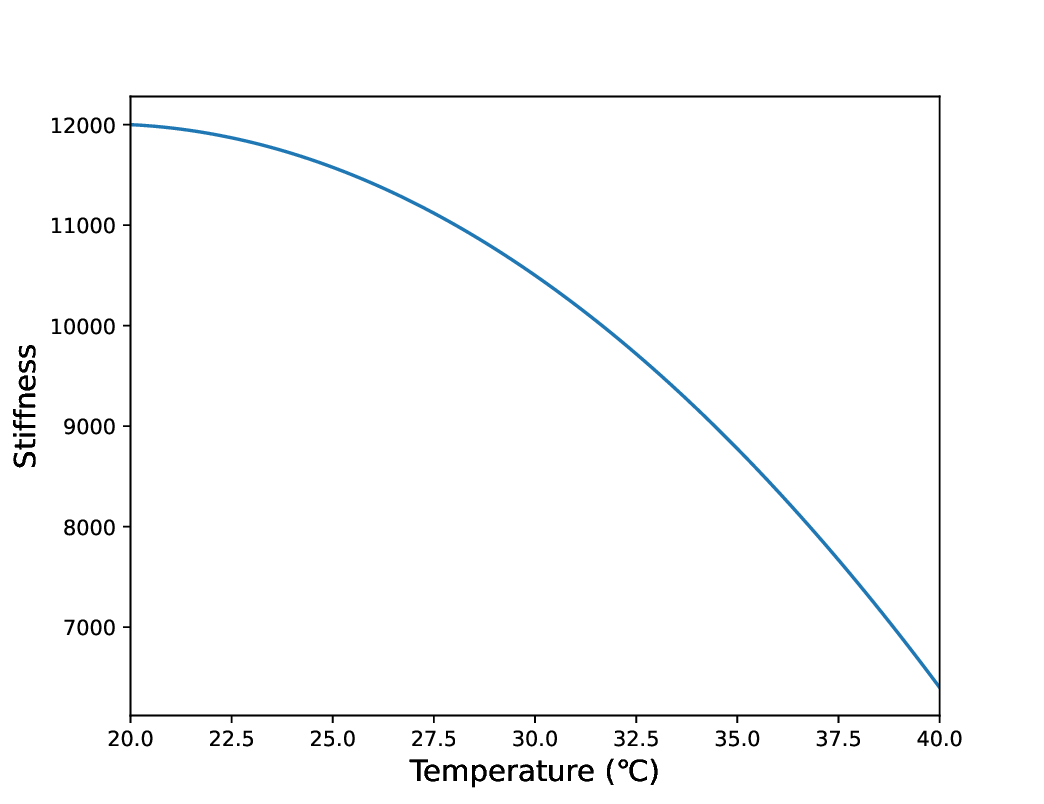}
    \caption{Relationship between the temperature and the stiffness of the first three springs of the systems of the population.}
    \label{fig:stiffness_temperature}
\end{figure}

Examples of \textit{frequency response functions} (FRFs) of the first degree of freedom of a system of the population are shown in Figure \ref{fig:varying_temp_FRFs}. The variation of the FRFs is shown as different colours, from lower temperatures (blue curves) to higher temperatures (red curves). The relationship of the stiffness parameter $k$ and the temperature is considered to be nonlinear and given by $k = -13T^{2} + 500T + 7200$ (Figure \ref{fig:stiffness_temperature}) and the temperature is considered to vary in the interval $[20, 40]$.

\begin{figure}[H]
\centering
\begin{subfigure}{.5\textwidth}
  \centering
    \includegraphics[width=0.3\paperwidth]{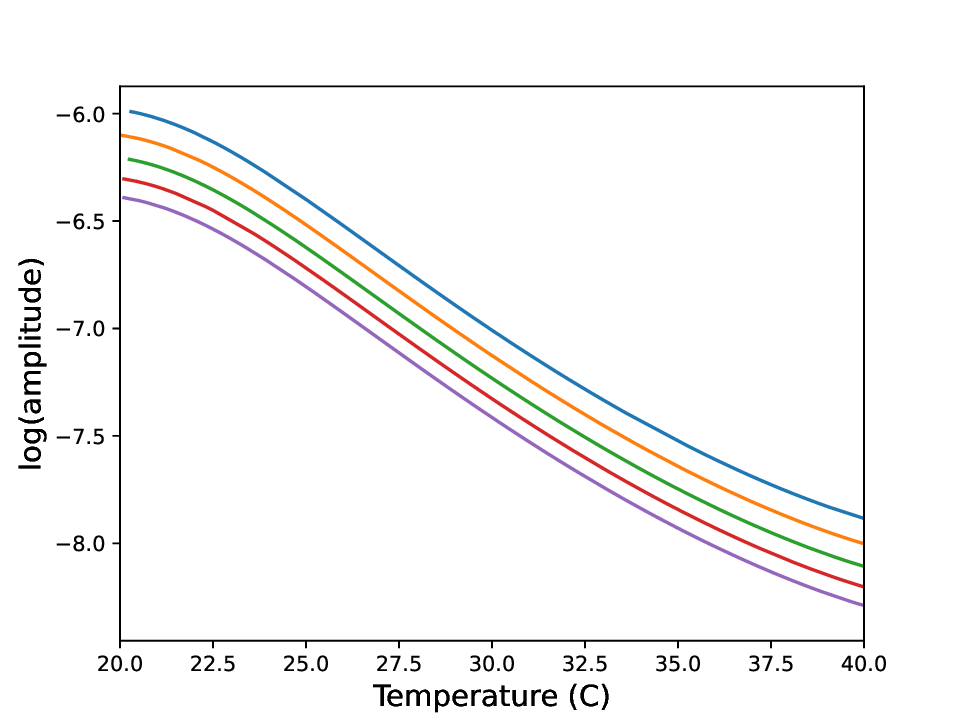}
    \label{fig:1_Hz}
\end{subfigure}%
\begin{subfigure}{.5\textwidth}
  \centering
    \includegraphics[width=0.3\paperwidth]{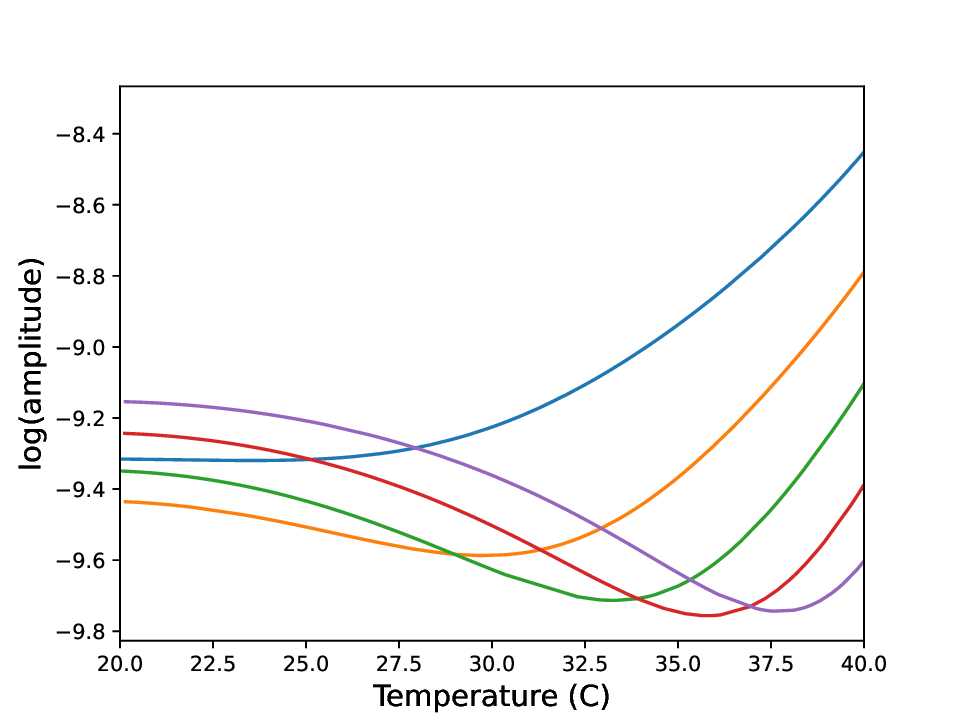}
    \label{fig:100_Hz}
\end{subfigure}
\caption{FRF line corresponding to frequency equal to 1 Hz (left) and 50 Hz (right), for values of the stiffness equal to $8000$ (blue curve), $9000$ (orange curve), $10000$ (green curve), $11000$ (red curve) and $12000$ (curve).}
\label{fig:spectral_lines_example}
\end{figure}

Three problems were studied in the current application. The first two refer to building neural network models that approximate the relationship between the temperature of the environment and the magnitude of a single spectral line from the FRFs. Two spectral lines were chosen for illustration, corresponding to $1$ Hz and $50$ Hz. Examples of the spectral lines for different temperatures and for three different structures are shown in Figure \ref{fig:spectral_lines_example}. The two problems are studied because of a characteristic which is clear from the two plots. In the first case, the $1$ Hz spectral lines, the relationship between the temperature and the task, and the value of the magnitude of the spectral line is a bijection. As a result, only one point suffices in order to characterise and identify the task for which inference is made; a functionality that both approaches should perform. In contrast, for the second case, overlapping between the curves is observed. Consequently it is expected that the latter shall be a more difficult task, because more than one point from each task is needed to perform the identification of the task.

\begin{figure}
    \centering
    \includegraphics[scale=0.5]{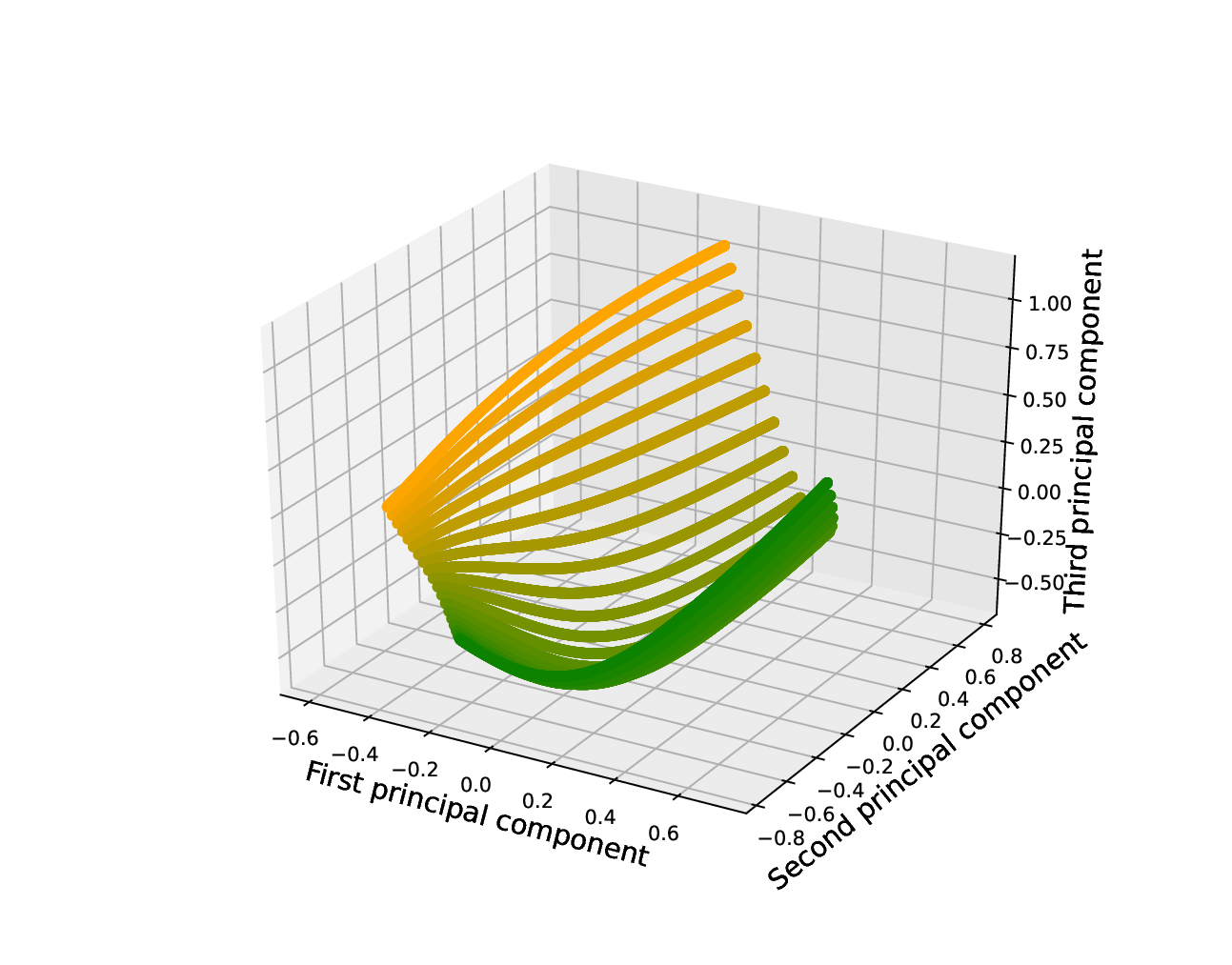}
    \caption{Principal components of samples of FRFs of members of the population. Points belonging to the same structure for different temperature are shown with the same colour. The colour difference refers to different structures of the population.}
    \label{fig:pca_pop_samples}
\end{figure}

The third problem is that of building a model to approach the whole FRF. To reduce the dimensionality of the FRFs, which are inferred, \textit{principal component analysis} (PCA) \cite{wold1987principal} was performed on the samples. For completeness, samples of the population for varying temperature are shown in Figure \ref{fig:pca_pop_samples}. In the aforementioned figure, one can see the effect of temperature on the modelled quantity. Moreover, the physics of the population data-wise are revealed. The transformation of the curves for different structures is shown. It is worth noting that the described algorithms are expected to imitate a human capability; observing the curves of Figure \ref{fig:pca_pop_samples}, one could infer the values of the corresponding curve of a structure, for which only a few points (even just one point in this case) are available. This human functionality is about understanding the physics of the data or, in the current case, the physics of the population.

The population-based set-up is defined, considering a small subset of structures as the \textit{training population}, for which data are freely available. Such a situation can be defined when one has extensively modelled a structure in operation or when one is able to test a structure in a laboratory for various excitations and environmental conditions. In contrast to other works on regression with meta-learning \cite{finn2017model, garnelo2018conditional, andrychowicz2016learning}, where tasks are freely sampled during training from a distribution and the models are trained essentially based on an infinite number of available tasks, in the current work the number of training tasks (structures) is considered to be small but data for these structures are available for a large number of different input variables. Using these data, neural network models are trained following the described approaches. Then, the models are used to approximate the relationship between the temperature and the FRF of new structures - the \textit{testing population} - for which only a few samples are available. The methods are then compared to a mainstream machine-learning approach, that of applying a \textit{Gaussian process} (GP) \cite{rasmussen2003gaussian} directly on the available data separately for every structure of the testing population. For the GP, a zero mean value function and a Gaussian kernel were selected. Different types of mean and kernel functions were not tested, because they are considered reasonable choices for a problem with a small amount of available data for which one has no prior intuition. The hyperparameters of the kernel function were optimised by maximising the likelihood of the model for the available data. 

\subsection{Application of MAML}

The exact procedure followed to train neural networks with the MAML method is similar to a train-validate-test approach of standard machine learning. In each case, a three-layer neural network is used, with one input layer, a hidden layer and an output layer. The activation function of the layers was a \textit{hyperbolic tangent} (tanh) function for all layers. For $N+1$ structures in the training population, Algorithm \ref{alg:MAML} is used to train a neural network based on the data of $N$ structures. The data from the structure which is left out and considered the \textit{validation structure}, are used to optimise the hyperparameters of the model; in the current case the size of the hidden layer of the neural network. The sizes of the hidden layer considered were from the set $\{10, 20..., 100 \}$. For each hidden-layer size, five random initialisations of the parameters of the network were considered and algorithm \ref{alg:MAML} was used to train the network. After every epoch of the algorithm, the model was adapted on the data of the validation structure; i.e.\ the model was trained by applying backpropagation using the data of the validation structure. The adapted model was tested on unseen-by-the-algorithm data from the validation structure. The instance of the model that achieved the lowest loss on the unseen data from the validation structure was considered the best model and was then tested on the structures from the testing population, whose number in all experiments was set equal to 200 structures. The whole procedure was repeated 50 times because of the random selection of the training population, which is expected to affect the performance of the model on the unseen structures of the population.

The training was repeated for different number of structures in the training population, as well as for different number of available samples for each testing structure. The model, which was previously trained using MAML, was adapted on samples coming from new testing structures and then the \textit{normalised mean-squared error} (NMSE) was calculated for 200 values of input temperatures randomly sampled from the interval $[20, 40]$. The NMSE of the performance of the algorithm is defined by,
\begin{equation}
    NMSE = \frac{100}{N \sigma_{y}^{2}}\sum_{i=1}^{N}(\hat{y}_{i} - y_{i})^{2}
    \label{eq:NMSE}
\end{equation}
where $\hat{y}_{i}$ is the prediction of the model for the $i$\textsuperscript{th} input sample, $y_{i}$ is the corresponding observation, $\sigma_{y}$ is the standard deviation of the values of the observations $y$ and $N$ is the total number of samples. This NMSE is useful as a metric since it is equal to $100\%$ if the model predictions ($\hat{y}_i$) are set to the mean value, i.e.\ $\hat{y}_i = \overline{y}$; values lower than $100\%$ reveal that the model is indeed capturing correlations in the data. Experience with this NMSE indicates that good models are obtained for values of less than 5\%, with a value of less than 1\% for excellent models. The results presented here refer to the average NMSE of the 50 repetitions of training and testing, as well as to the standard deviation of the 50 population NMSEs.

% The NMSEs are also analytically presented in Table \ref{Tab:NMSEs}.

% \begin{figure}
%     \centering
%     \includegraphics[scale=0.45]{Figures/NMSEs_barplot.eps}
%     \caption{Average normalised mean square error over a population of 200 testing structures, for models trained with different values of available structures in the training population and different number of available points for every testing structure.}
%     \label{fig:NMSEs_3D_barplot}
% \end{figure}

\begin{figure}[]
\centering
\begin{subfigure}{.5\textwidth}
  \centering
    \includegraphics[width=0.32\paperwidth]{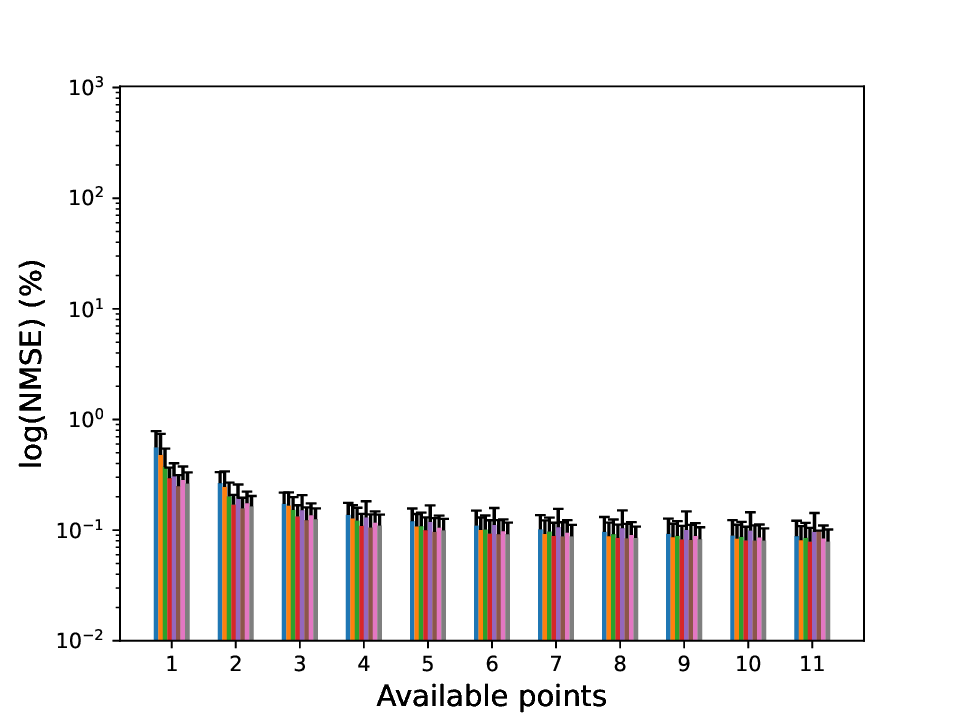}
    \label{fig:1_Hz_erbars_MAML}
\end{subfigure}%
\begin{subfigure}{.5\textwidth}
  \centering
    \includegraphics[width=0.32\paperwidth]{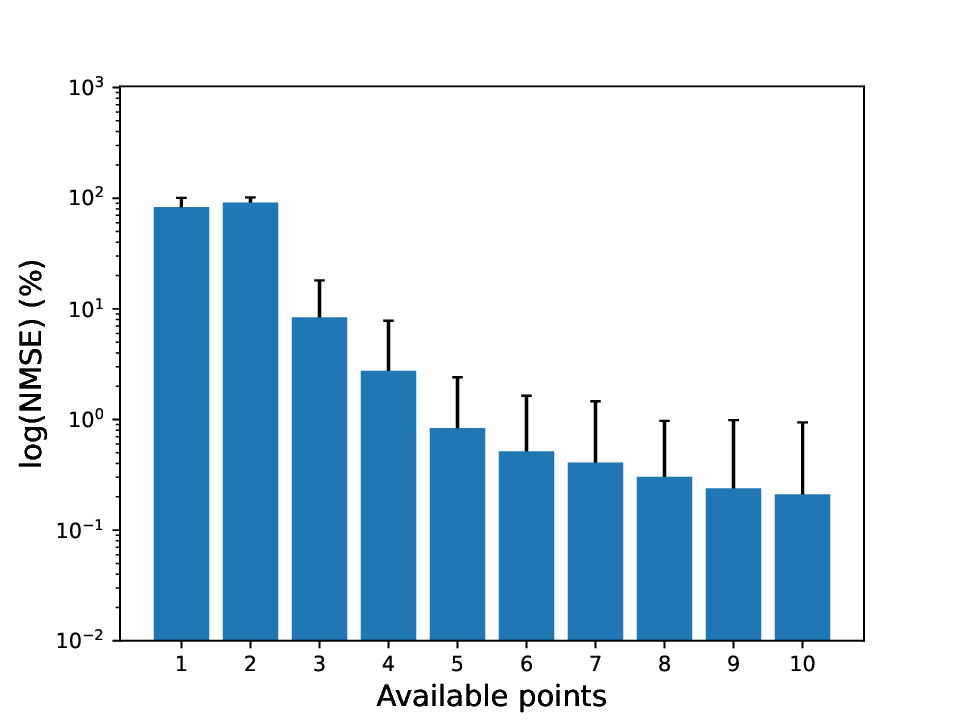}
    \label{fig:1_Hz_erbars_GP}
\end{subfigure}
\caption{Average normalised mean-squared errors and corresponding error-bars for the first problem, for a testing population of $200$ structures and for $100$ data samples for each structure using a neural network trained via the MAML algorithm (left) and for a GP (right). On the left, the different colours represent MAML-trained neural networks with a training population of two (blue), three (orange), four (green), five (red), six (purple), seven (brown), eight (pink) and nine (grey) structures.}
\label{fig:error_bars_1_Hz}
\end{figure}

\begin{figure}[]
\centering
\begin{subfigure}{.5\textwidth}
  \centering
    \includegraphics[width=0.32\paperwidth]{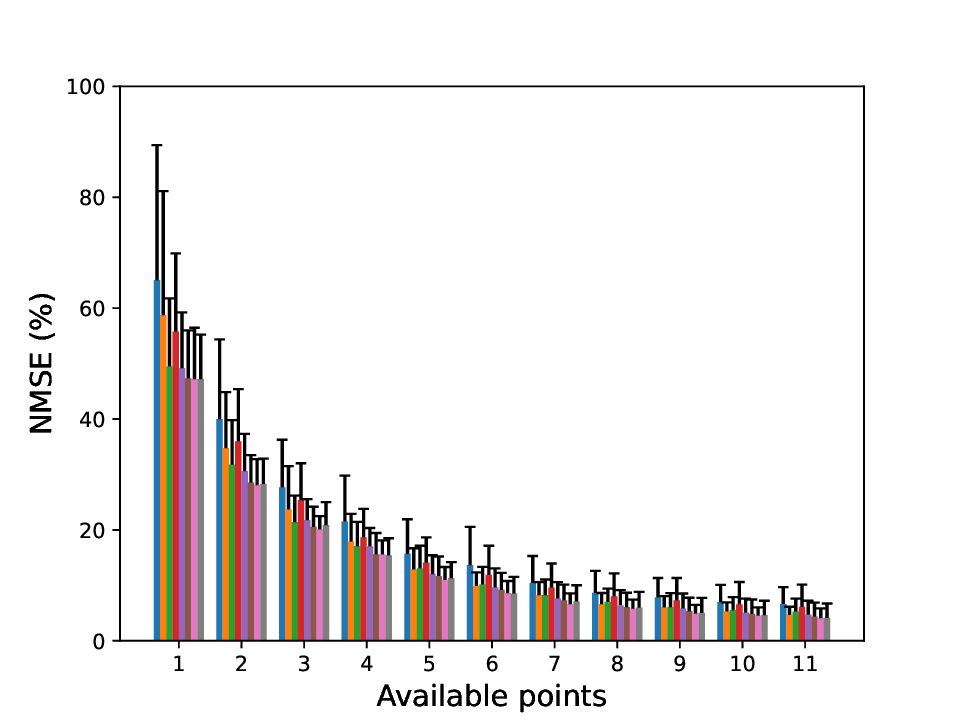}
    \label{fig:50_Hz_erbars_MAML}
\end{subfigure}%
\begin{subfigure}{.5\textwidth}
  \centering
    \includegraphics[width=0.32\paperwidth]{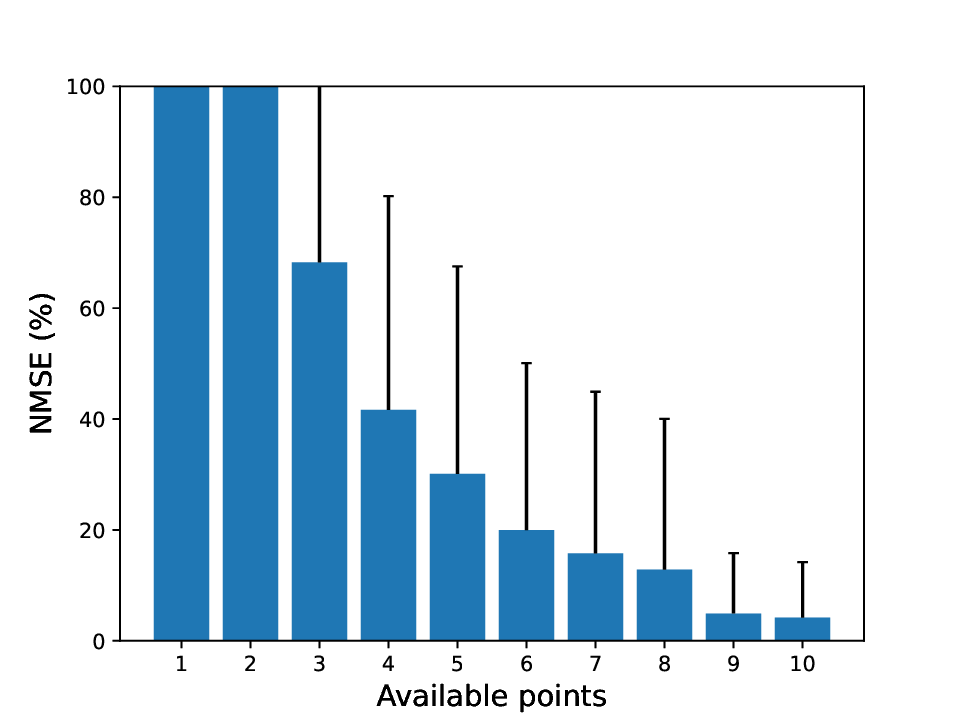}
    \label{fig:50_Hz_erbars_GP}
\end{subfigure}
\caption{Average normalised mean-squared errors and corresponding error-bars for the second problem, for a testing population of $200$ structures and for $100$ data samples for each structure using a neural network trained via the MAML algorithm (left) and for a GP (right). On the left, the different colours represent MAML-trained neural networks with a training population of two (blue), three (orange), four (green), five (red), six (purple), seven (brown), eight (pink) and nine (grey) structures.}
\label{fig:error_bars_50_Hz}
\end{figure}

\begin{figure}[]
\centering
\begin{subfigure}{.5\textwidth}
  \centering
    \includegraphics[width=0.32\paperwidth]{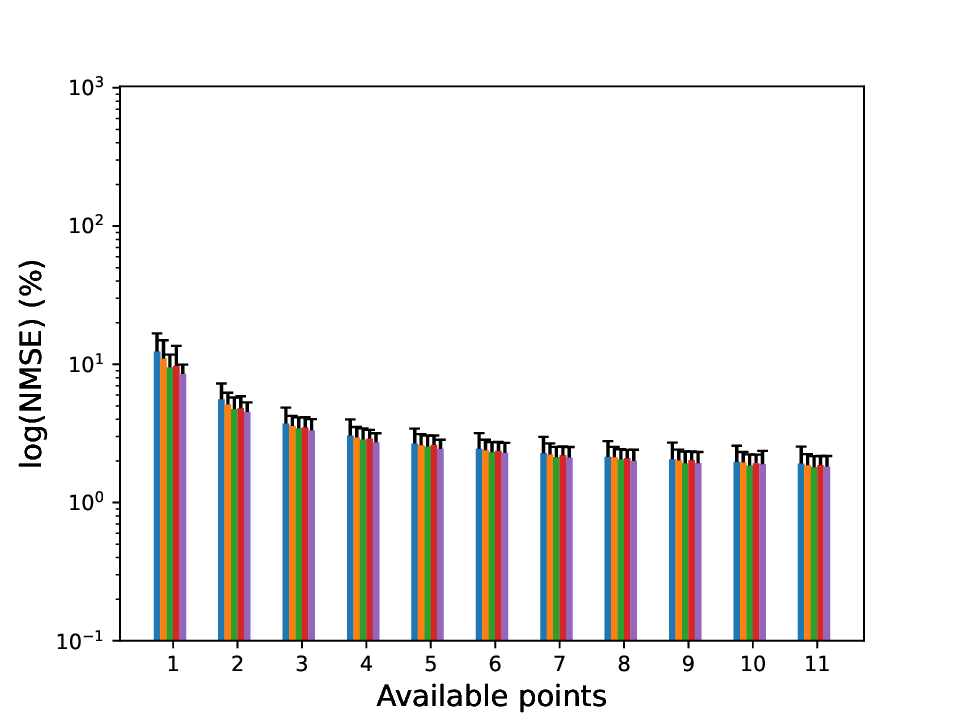}
    \label{fig:FRF_erbars_MAML}
\end{subfigure}%
\begin{subfigure}{.5\textwidth}
  \centering
    \includegraphics[width=0.32\paperwidth]{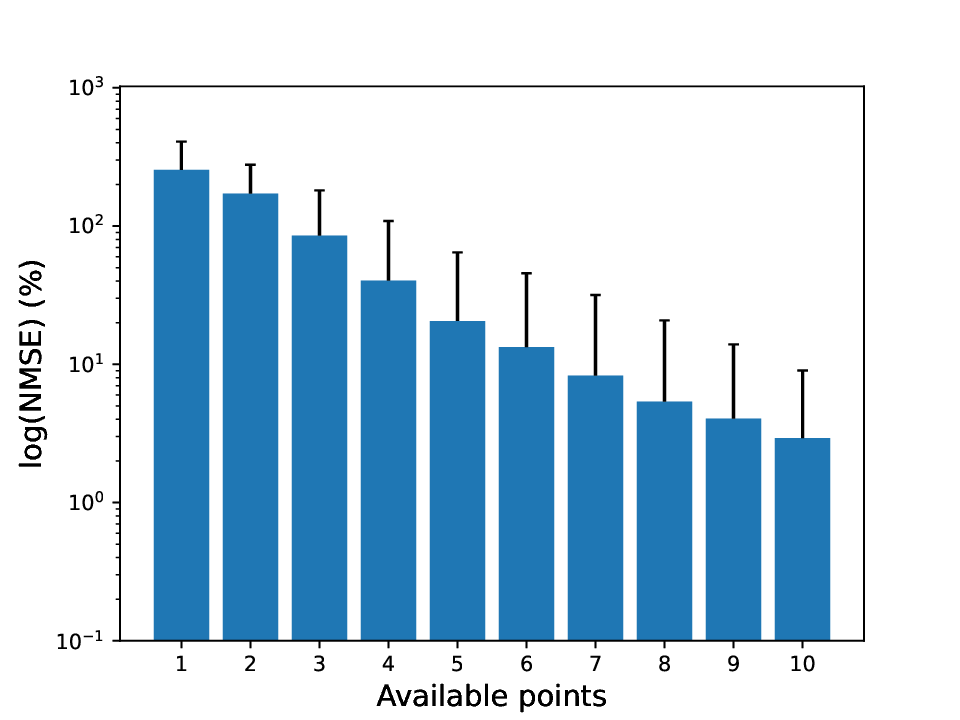}
    \label{fig:FRF_erbars_GP}
\end{subfigure}
\caption{Average normalised mean-squared errors and corresponding error-bars for the third problem, for a testing population of $200$ structures and for $100$ data samples for each structure using a neural network trained via the MAML algorithm (left) and for a GP (right). On the left, the different colours represent MAML-trained neural networks with a training population of two (blue), three (orange), four (green), five (red), six (purple), seven (brown), eight (pink) and nine (grey) structures.}
\label{fig:error_bars_FRF}
\end{figure}

The results for the three problems are presented in Figures \ref{fig:error_bars_1_Hz}, \ref{fig:error_bars_50_Hz} and \ref{fig:error_bars_FRF}; note that for the first and the third problems, the NMSE axis is shown on a logarithmic scale. As a first observation, one can easily see that the algorithm outperforms the GP approach in almost every case; this is to be expected, since the algorithm is trained based on data from the population, while the GP is not informed in any way by the population or by some prior physical knowledge of the analyst. However, for higher numbers of available training points, the GP is expected to achieve equal or lower NMSEs than the MAML algorithm, especially given that the presented problems are single-input single-output problems.

It can further be seen that the MAML algorithm functions as a traditional machine-learning algorithm regarding the number of available training data. The more available training structures, the lower the average error on the testing data and the lower the standard deviation of the errors in the population, indicating a more robust model. Small inconsistencies to these tendencies might be because of the random selection of the training structures, as well as the stochastic optimisation procedure of the neural-network models. This observation further encourages the belief that using the discussed algorithms may be a way to automatically extract exploitable knowledge from the population and use it to boost the performance of models for newly-presented structures. It is to the authors' knowledge that a GP can be informed by data from a population to boost its performance; however, such approaches may be included in the \textit{grey-box} modelling discipline \cite{cross2019grey, pitchforth2021grey}, according to which, one imposes knowledge into the model; such knowledge might also be acquired from a population of structures.

\begin{figure}
     \centering
     \begin{subfigure}[b]{0.48\textwidth}
         \centering
         \includegraphics[width=\textwidth]{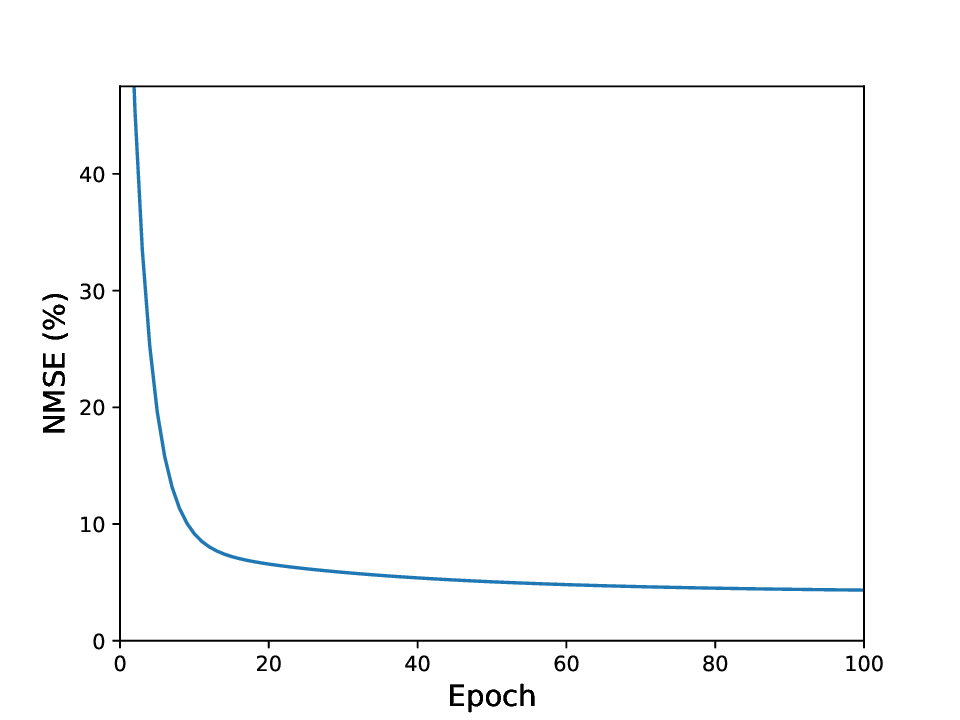}
         \caption{}
         \label{fig:converge_2}
     \end{subfigure}
     \hfill
     \begin{subfigure}[b]{0.48\textwidth}
         \centering
         \includegraphics[width=\textwidth]{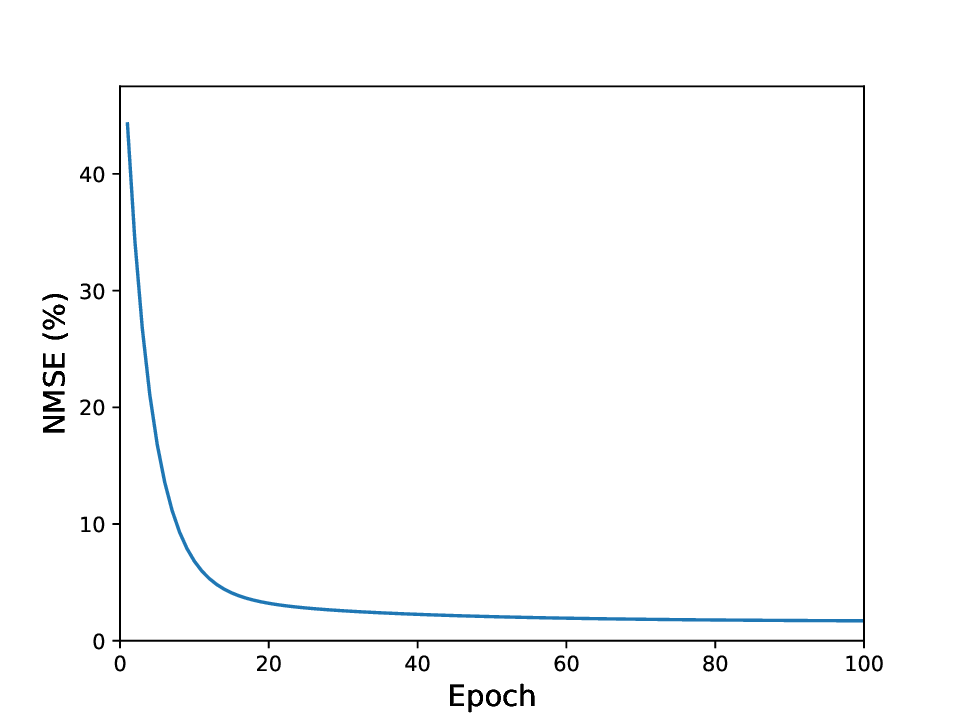}
         \caption{}
         \label{fig:converge_3}
     \end{subfigure}
        \caption{Average normalised mean-square error history examples from the second problem, for (a) five training structures and five available samples for the testing structure and (b) nine training structures and seven available samples for the testing structure.}
        \label{fig:convergence}
\end{figure}

It is also worth examining the results even further; specifically the convergence of the neural-network model, which was trained using MAML. As mentioned in the original work, the model which is trained using MAML, is quite sensitive to different tasks. In the current case, different tasks refer to different structures. The average NMSE history for different sizes of the training population and available samples for the structures of the testing population are shown in Figure \ref{fig:convergence}. The average NMSE refers to the 200 testing points of values of temperature uniformly sampled from the interval $[20, 40]$. It is clear that as the number of structures in the training population and the number of available samples for the testing structures increase, so does the convergence rate towards a minimum. This proportionality indicates that larger sizes of training populations allow the algorithm to learn better the physics of the population. Another important aspect of the training histories, is how stable they are. After some epochs, a minimum has been achieved and, even though the model is trained for longer than that, the error does not diverge significantly from the achieved minimum. The observed stability in the NMSE training histories also encourages the belief of the previous section, that using MAML, the trainable parameters of the model are ``trapped'' on a manifold, where the solutions for the population exist. In contrast to what is observed in the current figures, a model, whose trainable parameters would be freely allowed to adapt, would potentially start to overfit to the values of the training data and would not exhibit such a stable training history regarding the error on the testing data.

\subsection{Application of CNP}

\begin{figure}[]
\centering
\begin{subfigure}{.5\textwidth}
  \centering
    \includegraphics[width=0.32\paperwidth]{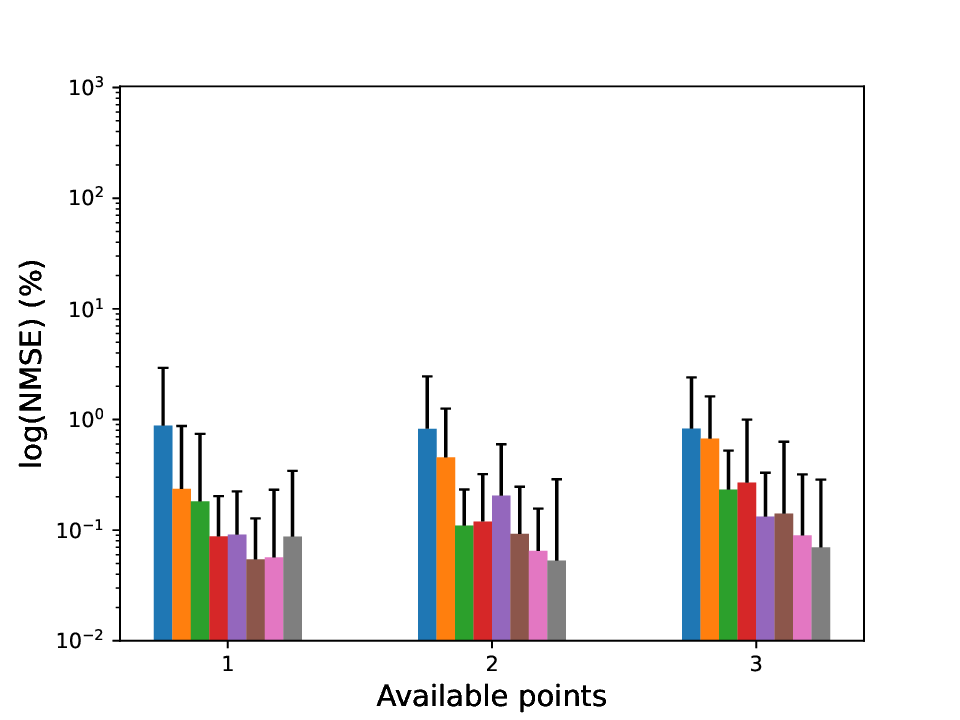}
    \label{fig:1_Hz_erbars_CNP}
\end{subfigure}%
\begin{subfigure}{.5\textwidth}
  \centering
    \includegraphics[width=0.32\paperwidth]{Figures/GP_1_Hz.eps}
    \label{fig:1_Hz_erbars_GP_2}
\end{subfigure}
\caption{Average normalised mean-squared errors and corresponding error-bars for the first problem, for a testing population of $200$ structures and for $100$ data samples for each structure using a neural network trained via the CNP algorithm (left) and for a GP (right). On the left, the different colours represent CNP-trained neural networks with a training population of two (blue), three (orange), four (green), five (red), six (purple), seven (brown), eight (pink) and nine (grey) structures.}
\label{fig:error_bars_1_Hz_CNP}
\end{figure}

\begin{figure}[]
\centering
\begin{subfigure}{.5\textwidth}
  \centering
    \includegraphics[width=0.32\paperwidth]{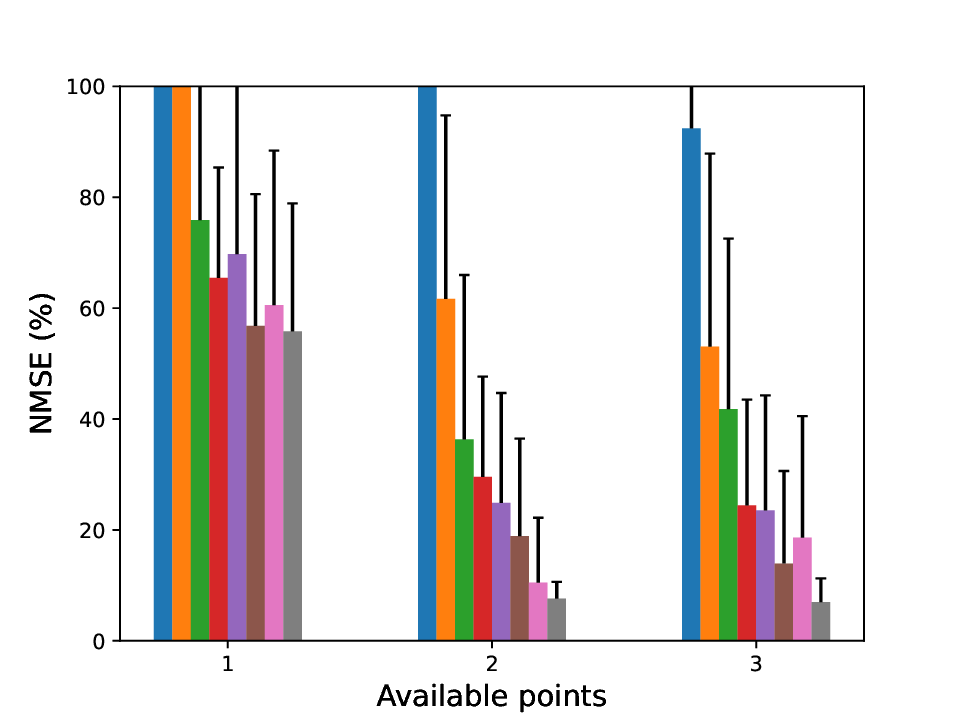}
    \label{fig:50_Hz_erbars_CNP}
\end{subfigure}%
\begin{subfigure}{.5\textwidth}
  \centering
    \includegraphics[width=0.32\paperwidth]{Figures/GP_50_Hz.eps}
    \label{fig:50_Hz_erbars_GP_2}
\end{subfigure}
\caption{Average normalised mean-squared errors and corresponding error-bars for the second problem, for a testing population of $200$ structures and for $100$ data samples for each structure using a neural network trained via the CNP algorithm (left) and for a GP (right). On the left, the different colours represent CNP-trained neural networks with a training population of two (blue), three (orange), four (green), five (red), six (purple), seven (brown), eight (pink) and nine (grey) structures.}
\label{fig:error_bars_50_Hz_CNP}
\end{figure}

\begin{figure}[]
\centering
\begin{subfigure}{.5\textwidth}
  \centering
    \includegraphics[width=0.32\paperwidth]{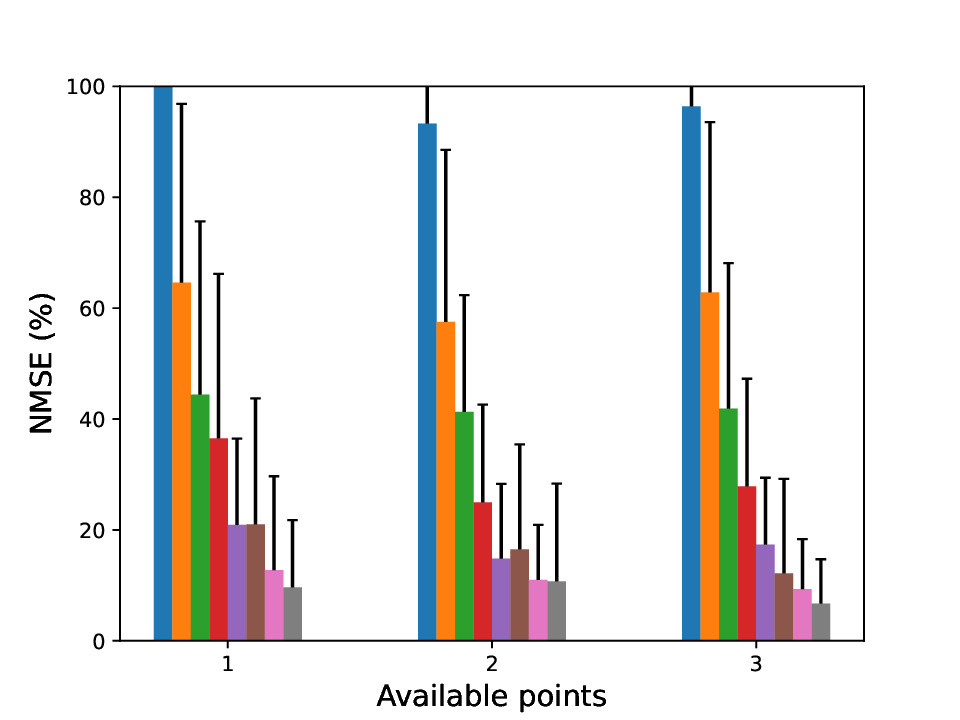}
    \label{fig:FRF_erbars_CNP}
\end{subfigure}%
\begin{subfigure}{.5\textwidth}
  \centering
    \includegraphics[width=0.32\paperwidth]{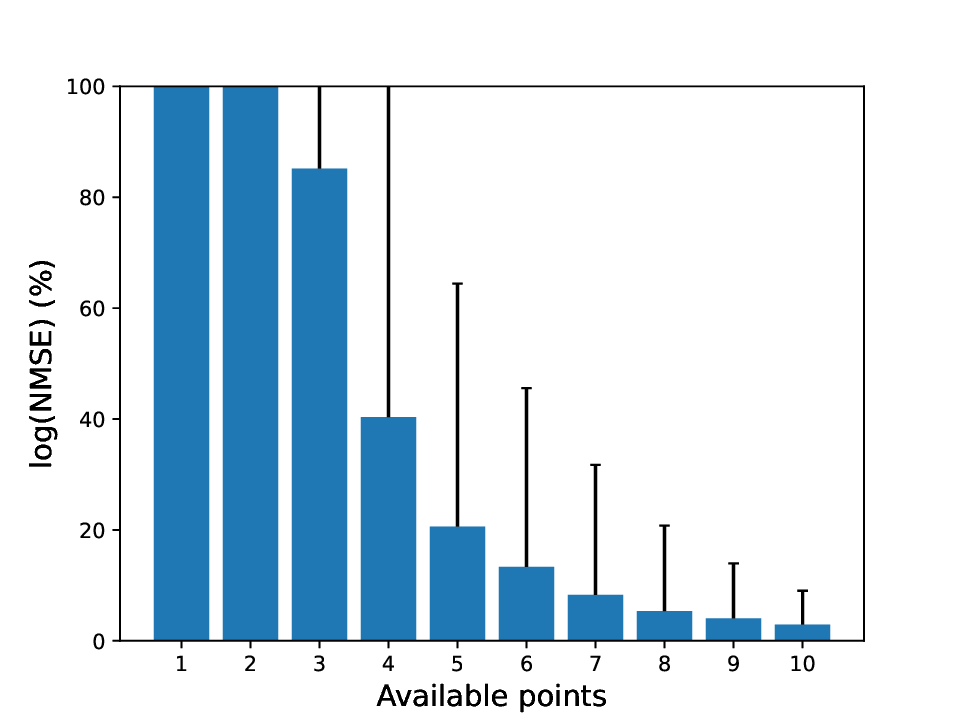}
    \label{fig:FRF_erbars_GP_2}
\end{subfigure}
\caption{Average normalised mean-squared errors and corresponding error-bars for the third problem, for a testing population of $200$ structures and for $100$ data samples for each structure using a neural network trained via the CNP algorithm (left) and for a GP (right). On the left, the different colours represent CNP-trained neural networks with a training population of two (blue), three (orange), four (green), five (red), six (purple), seven (brown), eight (pink) and nine (grey) structures.}
\label{fig:error_bars_FRF_CNP}
\end{figure}

The same framework was followed to test the performance of a CNP model on the population. Results are presented in a similar manner in Figures \ref{fig:error_bars_1_Hz_CNP}, \ref{fig:error_bars_50_Hz_CNP} and \ref{fig:error_bars_FRF_CNP}. As in the case of applying MAML, the algorithm also exhibits behaviour similar to traditional machine learning. As the number of available training structures increases, the error almost monotonically decreases. Similar to before, some inconsistencies are observed, which might be caused by the random selection of the training populations. The same behaviour is observed for the standard deviation of the error, which also decreases as more training structures are available, rendering a more robust model for the population.

The CNP model also appears to perform much better than the GP in the case of problems one and three, which, as discussed, is expected. It is also observed that for the second problem, the CNP approach performs better than MAML when more training structures are available. The better performance in such cases might be because of the different ways that the two models are trained. The MAML approach is an attempt to find an initialisation point for the trainable parameters of the neural network and from there to quickly fit a model to the task-specific optimal parameters. MAML is therefore an implicit way of identifying the task via the error signals passed by the backpropagation process to the model. On the contrary, CNP is optimised exclusively with a view to identifying the task from the available data, inferring a task-descriptive vector and using this vector as a feature to make predictions for new inputs. It appears that MAML, where backpropagation is performed through the optimisation algorithm, is a more complicated attempt to learn the physics of the population and, as a result, the more direct approach of CNP in this case outperforms MAML.

% \begin{figure}[]
% \centering
% \begin{subfigure}{.5\textwidth}
%   \centering
%     \includegraphics[width=0.32\paperwidth]{Figures/GP_1_Hz_conf_intervals.pdf}
%     \label{fig:fit_example_GP_1_Hz}
% \end{subfigure}%
% \begin{subfigure}{.5\textwidth}
%   \centering
%     \includegraphics[width=0.32\paperwidth]{Figures/GP_50_Hz_conf_intervals.pdf}
%     \label{fig:fit_example_GP_50_Hz}
% \end{subfigure}
% \caption{The mean of the predictions (orange curve) and confidence intervals of $\pm 3$ standard deviations of the predictions (shaded blue area) of a GP fitted to available data from a testing structure (red star points) for the 1 Hz problem (left) and the 50 Hz problem (right). The real underlying relationship is shown as the blue curve and the predictions of the model as the orange curve.}
% \label{fig:fit_example_GP}
% \end{figure}

\begin{figure}[]
\centering
\begin{subfigure}{.5\textwidth}
  \centering
    \includegraphics[width=0.32\paperwidth]{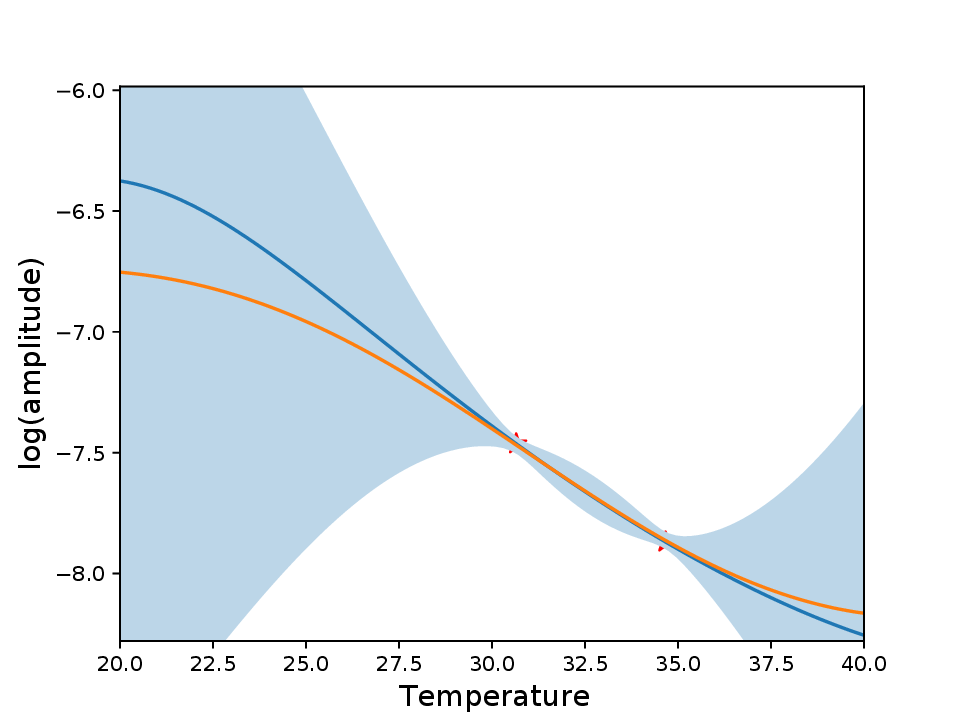}
    \label{fig:fit_example_GP_1_Hz}
\end{subfigure}%
\begin{subfigure}{.5\textwidth}
  \centering
    \includegraphics[width=0.32\paperwidth]{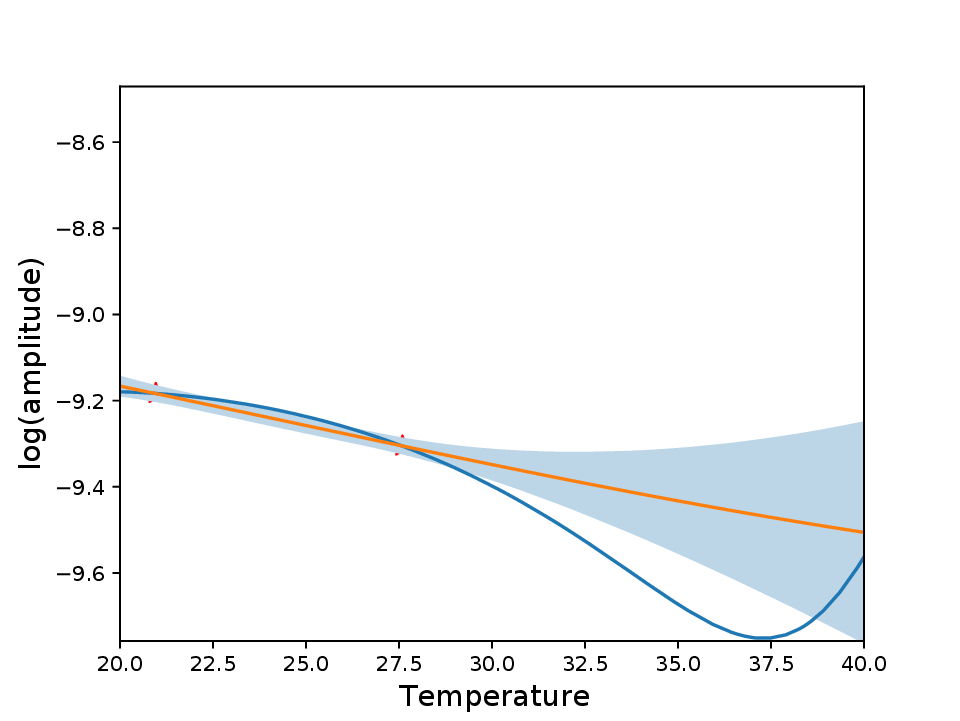}
    \label{fig:fit_example_GP_50_Hz}
\end{subfigure}
\caption{The mean of the predictions (orange curve) and confidence intervals of $\pm 3$ standard deviations of the predictions (shaded blue area) of a GP fitted to available data from a testing structure (red star points) for the 1 Hz problem (left) and the 50 Hz problem (right). The real underlying relationship is shown as the blue curve and the predictions of the model as the orange curve.}
\label{fig:fit_example_GP}
\end{figure}

\begin{figure}[]
\centering
\begin{subfigure}{.5\textwidth}
  \centering
    \includegraphics[width=0.32\paperwidth]{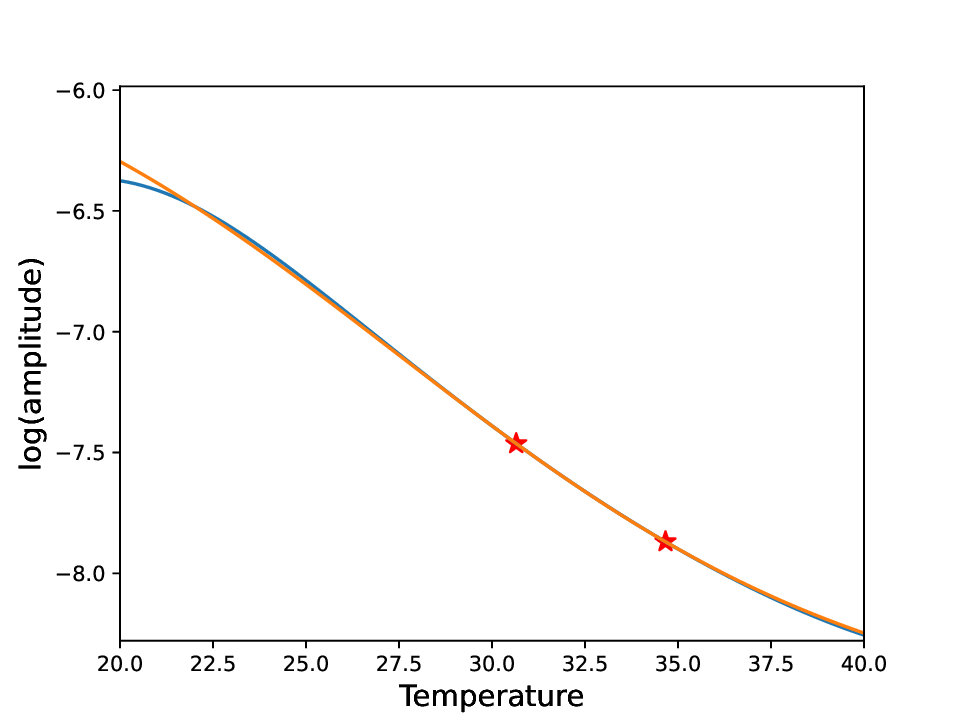}
    \label{fig:fit_example_MAML_1_Hz}
\end{subfigure}%
\begin{subfigure}{.5\textwidth}
  \centering
    \includegraphics[width=0.32\paperwidth]{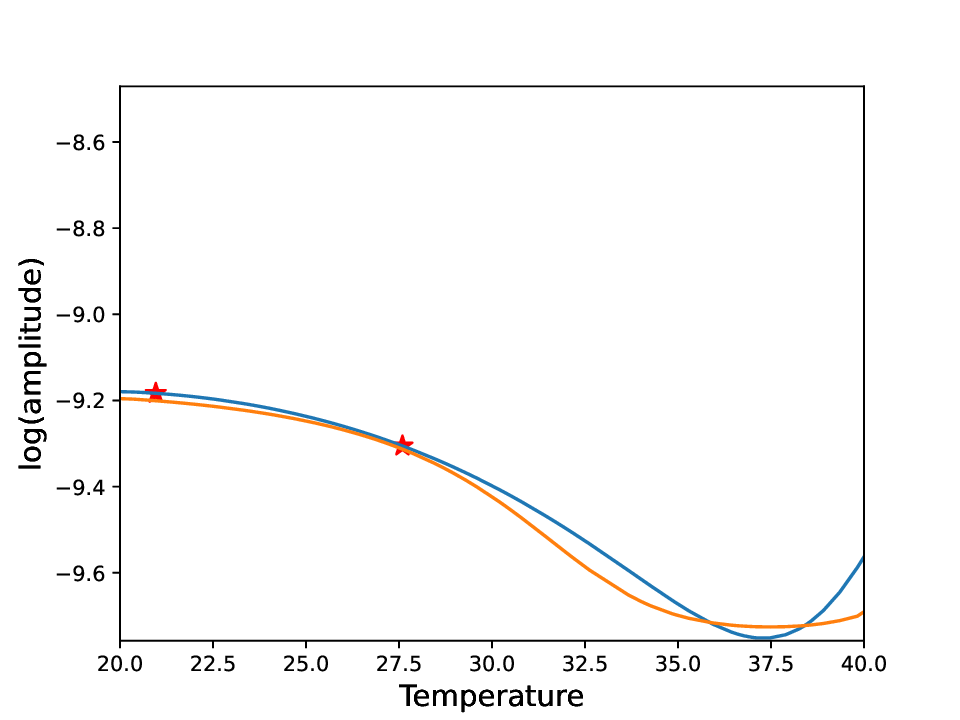}
    \label{fig:fit_example_MAML_50_Hz}
\end{subfigure}
\caption{Example of fitting a neural network, trained according to the MAML algorithm on nine training structures, to available data from a testing structure (red star points) for the 1 Hz problem (left) and the 50 Hz problem (right). The real underlying relationship is shown as the blue curve and the predictions of the model as the orange curve.}
\label{fig:fit_example_MAML}
\end{figure}

\begin{figure}[]
\centering
\begin{subfigure}{.5\textwidth}
  \centering
    \includegraphics[width=0.32\paperwidth]{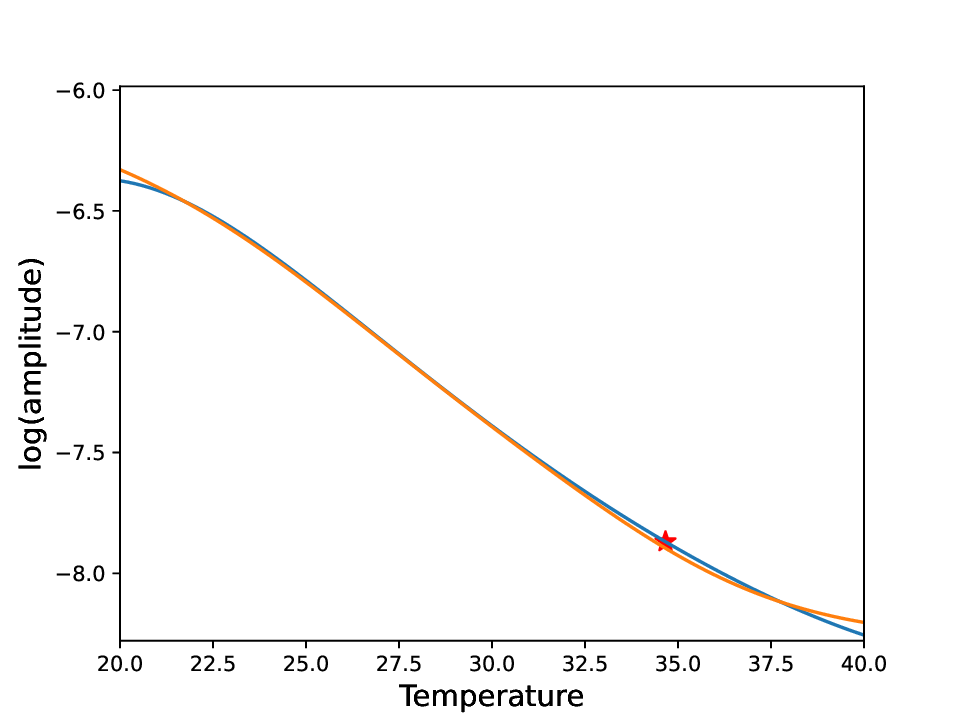}
    \label{fig:fit_example_CNP_1_Hz}
\end{subfigure}%
\begin{subfigure}{.5\textwidth}
  \centering
    \includegraphics[width=0.32\paperwidth]{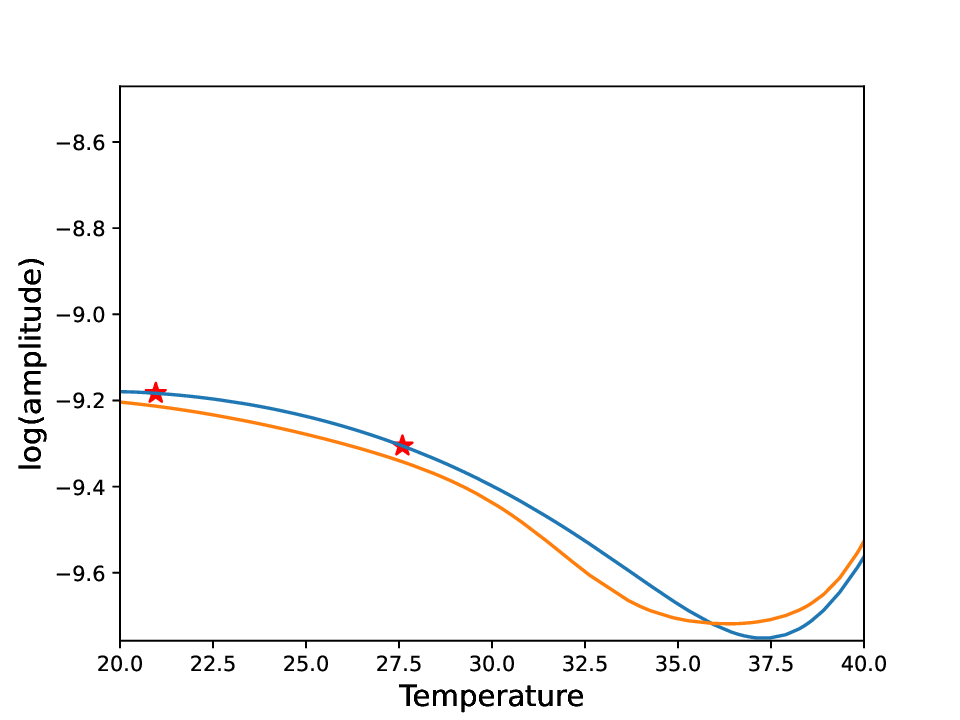}
    \label{fig:fit_example_CNP_50_Hz}
\end{subfigure}
\caption{Example of the performance of a CNP model, trained using data from nine training structures. The algorithm uses data from testing structure (red star points) to make predictions for the 1 Hz problem (left) and the 50 Hz problem (right). The real underlying relationship is shown as the blue curve and the predictions of the model as the orange curve.}
\label{fig:fit_example_CNP}
\end{figure}

To better understand why population-informed models outperform the traditional machine-learning approach, some examples of fitting these models to a small subset of data from a testing structure are shown in Figures \ref{fig:fit_example_GP}, \ref{fig:fit_example_MAML} and \ref{fig:fit_example_CNP}. For the current examples, the population-informed models are trained using a training population of nine structures. In the figures, examples of the performance of the models for the first two problems are presented. Observing the behaviour of the population-informed models away from the available samples gives a clear indication that these models have incorporated part of the physics of the structures. For CNP it is also clear that with only one available sample from the testing structure, the model is able to almost perfectly approximate the underlying relationship, although, as discussed the first problem is quite a simple problem. The GP on the other hand is not informed in any way from the population. As a result, its predictions are not as efficient as the predictions of the other two models.

\section{Conclusions and next steps}
\label{sec:conclusions}

The current work aims at motivating the creation of data-driven models, which are forced to respect the underlying physics of some population of problems. The physics of the population are to be learnt via the use of data from a population of structures. The desired result is inspired by the functionality of physics-based models, which utilise parameters that describe the characteristics of various structures. These parameters belong to a predefined interval, in contrast to the trainable parameters of a data-driven model, whose domain is the whole set of real numbers.

Two approaches are studied for the purpose of defining population-informed models. The first is based on forcing the trainable parameters of a neural network model to lie on a manifold, where minima for modelling structures of the population exist. The approach is implemented using a meta-learning technique for training neural networks, the model-agnostic meta-learning algorithm (MAML). The second approach is based on using a set of available input-output observations from a structure to identify the structure for which inference is to be made. The approach is implemented via the conditional neural processes algorithm (CNP), which exploits a task-identification subnetwork to infer a task-descriptive vector, from a small subset of available task-specific data, and use this vector as a feature in the main neural network model to make predictions about input-variables of interest.

The two types of population-informed models are tested based on three simulated datasets. The datasets comprise FRFs of simulated lumped-mass structures as a function of the temperature of the environment of the structures. The two first datasets refer to the prediction of the magnitude of single spectral lines of the FRF of the lumped-mass systems and the third problem is about inferring the whole FRF. The difference between the first two problems is that for the first, a one-to-one relationship exists between the structure (task) and the temperature, and the magnitude of the spectral line, while for the second task overlapping between the task curves exists.

The algorithms in all cases are tested on a small population of available structures, in contrast to other works on meta-learning where one had unrestricted access to random tasks during training. Moreover, because the selection of the training structures naturally affects the result, the training and testing of the algorithms were performed several times and the average NMSE and the standard deviation of the NMSEs were calculated. The performance of the two algorithms was compared to a traditional machine-learning method - a Gaussian process (GP) - which is suitable for cases of modelling datasets with only a few available data.

The results reveal that both algorithms are able to exploit knowledge from the population. The algorithms are able to perform better than the GP, which, in the current work, is not informed by the population or prior knowledge of the analysers. Both MAML and CNP exhibit low average population errors for the second case study and very low average population errors for the first and third problems, where the relationship between the temperature and the structure, and the quantity of interest is a bijection. Moreover, both algorithms seem to behave similarly to a traditional machine-learning algorithm regarding the number of available training structures and the average error across the testing population as well as the standard deviation of these errors. It is clear that the more training structures one has, the more accurate and robust the model is within the population. It is also worth noting, that the current approach is a completely physics-blind approach, in the sense that one does not need to induce any kind of physical knowledge in the algorithm for them to function properly. In future work, it would be interesting to examine a combined approach, including some physics into the procedure.

Although it is not the algorithms' primary objective, they could potentially be used to identify damaged structures. Including data from structures with different levels of damage in the population, the algorithm could learn to make inferences for such situations. Thus, with a few data from a structure, one could infer its behaviour for a wider range of environmental or operational conditions. Making such inferences, the comparison between the healthy states and the testing states could be made more efficiently. 

The algorithms are tested only on simulated data. However, the current work aims at motivating the use of black-box models in a similar manner to physics-based models. A major ability of physics-based models, is their reusability. The models can be built for a specific structure and then slightly modified to be used on another. To bridge the gap between the use of the two types of models, the approach herein is to define neural network models which are built in order to perform inference within a population. The two presented methods appear to create models suitable for such applications. Further validation of the methods on experimental data is needed; however, the results are encouraging. It is also believed that such models may add extra credibility to the use of data-driven models, because their behaviour and their goal is better defined in terms of being models of a population, rather than models that are only able to perform on a single task. Models defined in the described ways could more easily be trusted, as they can be validated for a set of structures of the population. Because they draw information from a population, the models do not begin their learning from complete ignorance of the underlying physics of the population, which renders them as not completely physics-ignorant models and makes their behaviour more trustworthy than traditional machine-learning methods.

% Exploiting data from a population, a definition of a small subset of trainable parameters of the black-box model is proposed. The functionality of this subset of trainable parameters is to imitate the functionality of the structural parameters which are used when one attempts to model a structure using a physics-based model. By achieving a definition of such a manifold, one could be confident that using parameters from the manifold, the model can sufficiently perform inference for the various members of the population. 

% To achieve such a restriction of parameters of a neural network model, in the current work, the MAML training strategy was used. The framework is tested on a simulated population of structures with varying initial structural parameters, which are also affected by environmental conditions, in this case, temperature. A neural network model is trained using different sizes of training populations and different numbers of available samples for adapting to new testing structures. The results reveal that the model achieves quite high average accuracy within a testing population, even for very small numbers of training structures and available testing samples. The accuracy is compared with the accuracy of a standard Gaussian process approach, which does not make use of any information from the population. The population-based approach yielded significantly better results, which was expected especially in the case of very few available samples for each testing structure. 

\section*{Acknowledgements}
\label{sec:ack}

The authors wish to gratefully acknowledge support for this work through grants from the Engineering and Physical Sciences Research Council (EPSRC), UK, via the Programme Grant EP/R006768/. For the purpose of open access, the authors have applied a Creative Commons Attribution (CC BY) licence to any Author Accepted Manuscript version arising.

% References
\bibliographystyle{unsrt}
\bibliography{meta_learning}

\end{document}